%% file: main_arxiv.tex
\documentclass[10pt,twocolumn,letterpaper]{article}

\usepackage{cvpr}              % To produce the CAMERA-READY version
\input{preamble}

\definecolor{cvprblue}{rgb}{0.21,0.49,0.74}
\usepackage[pagebackref,breaklinks,colorlinks,citecolor=cvprblue]{hyperref}

\usepackage{graphicx}
\usepackage{amsmath}
\usepackage{amssymb}
\usepackage{booktabs}
\usepackage{float}
\usepackage{stfloats}
\usepackage{multirow}
\usepackage{diagbox}
\usepackage{makecell}
\usepackage[toc,page]{appendix}

\usepackage{algorithm}
\usepackage{algpseudocode}
 
\usepackage[accsupp]{axessibility}

\usepackage[capitalize]{cleveref}
\crefname{section}{Sec.}{Secs.}
\Crefname{section}{Section}{Sections}
\Crefname{table}{Table}{Tables}
\crefname{table}{Tab.}{Tabs.}

\def\paperID{7293} % *** Enter the Paper ID here
\def\confName{CVPR}
\def\confYear{2024}

\title{Neural Video Compression with Feature Modulation}

\begin{document}

\author{Jiahao Li, Bin Li, Yan Lu\\
Microsoft Research Asia\\
{\tt\small \{li.jiahao, libin, yanlu\}@microsoft.com}
}
\maketitle

\begin{abstract}
 The emerging conditional coding-based neural video codec (NVC) shows superiority over commonly-used residual coding-based codec and the latest NVC already claims to outperform the best traditional codec. However, there still exist critical problems blocking the practicality of NVC. In this paper, we propose a  powerful conditional coding-based NVC that solves two critical problems via feature modulation. The first is how to support a wide quality range in a single model. Previous NVC with this capability only supports about 3.8 dB PSNR range on average. To tackle this limitation, we modulate the latent feature of the current frame via the learnable quantization scaler.
 During the training, we specially design the uniform quantization parameter sampling mechanism to improve the harmonization of encoding and quantization. This results in a better learning of the quantization scaler and helps our NVC support about 11.4 dB PSNR range. The second is how to make NVC still work under a long prediction chain. We expose that the previous SOTA NVC has an obvious quality degradation problem when using a large intra-period setting. To this end, we propose modulating the temporal feature with a periodically refreshing mechanism to boost the quality. %Besides solving the above two problems, we also design a single model that can support both RGB and YUV colorspaces. 
 Notably, under single intra-frame setting, our codec can achieve 29.7\% bitrate saving over previous SOTA NVC with 16\% MACs reduction. Our codec serves as a notable landmark in the journey of NVC evolution. The codes are at \url{https://github.com/microsoft/DCVC}.

\end{abstract}

\section{Introduction}
\label{sec:intro}
 %Although the prototype of next-generation traditional codec/ECM currently can save about 18\% bitrate over H.266/VTM \cite{bross2021overview}, the encoding complexity is increased to more than 6 times \cite{ECM_performance2023}.

%As the volume of video data expands at an unprecedented rate, video compression has become increasingly vital for storage and transmission. 
Traditional standard codec, which relies on a hybrid residual coding-based framework, has been in development for over 30 years and is still being refined. However, the improvement in compression ratio has diminished, while the increase in complexity has grown significantly \cite{ECM_performance2023}. This makes further advancements within the traditional framework more and more challenging. Recently, neural video codec (NVC) has gained considerable attention, as it holds the potential to break this development bottleneck.

%Recent years have witnessed the prosperity of NVC.
The early NVC model  DVC \cite{lu2019dvc}  still follows traditional codec and uses the residual coding-based framework. %, where all sub-modules are replaced by the neural network to achieve end-to-end learning.
Later, many works \cite{hu2022coarse,lin2020m,agustsson2020scale, liu2023mmvc} are also based on this paradigm and propose stronger sub-modules to improve the performance. By contrast, the emerging conditional coding  \cite{ladune2020optical, li2021deep} shows a lower entropy bound than residual coding and has larger potential. The condition can be freely defined and learned rather than being limited to the predicted frame in pixel domain. Meanwhile, the condition can be flexibly used to help encoding, decoding, and entropy modelling. The recent  DCVC-DC model \cite{li2023neural} already achieves a better compression ratio than both H.266/VTM \cite{bross2021overview} and the under-development ECM (the prototype of next generation traditional standard) by mining diverse spatial and temporal contexts as the condition.

\begin{figure}[t]
		\begin{center}
			\includegraphics[width=1.02\linewidth]{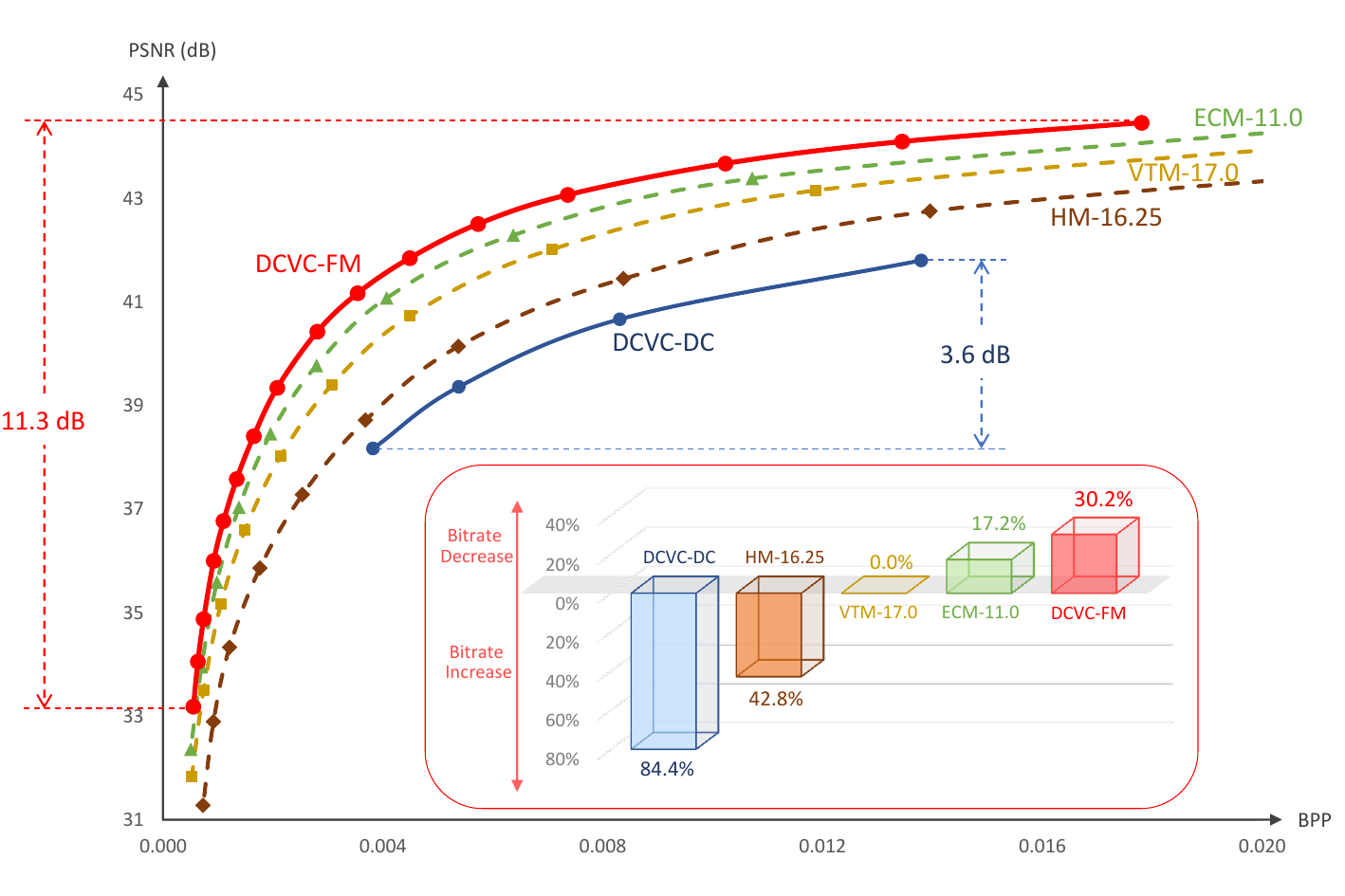}
		\end{center}
		\vspace{-0.7cm}
		\caption{Rate-Distortion curve and BD-Rate comparisons with H.265/HM, H.266/VTM, ECM, and the previous SOTA NVC DCVC-DC \cite{li2023neural}. The test dataset is HEVC E  (600 frames) with single intra-frame setting (i.e. intra-period = --1) and YUV420 colorspace. DCVC-DC has a large performance drop under this setting while our new codec DCVC-FM still can significantly surpass ECM. Meanwhile, the quality range of our DCVC-FM is much larger than that of DCVC-DC.	}
		\vspace{-6mm}
		\label{first_page}
	\end{figure}

Despite the progress made by DCVC-DC, we find it is still far from practical usage. The first blocking issue is lacking the wide quality range support. Although DCVC-DC can support multiple quality levels in a single model, its quality range is quite limited and only has an average of 3.8 dB on various datasets, which definitely cannot meet the needs of a practical product. To solve this problem, we modulate the latent feature of the current frame with the learnable quantization scaler. During the training phase, we not only increase the lambda range, but also specially design a uniform quantization parameter sampling mechanism to let NVC experience various trade-offs between rate and distortion.
This can enhance the harmonization between encoding and quantization processes. Consequently, the finely-controllable quantization scaler is obtained and our NVC can seamlessly adjust the quality level under a wide quality range, i.e. about 11.4 dB. 
Furthermore, with the support of a wide quality range, this paper also showcases the capability of rate control  in a single model, when given the specified target bitrate. This effectively demonstrates the practicality of our NVC in real-world scenarios.

The second issue is how to enable NVC to cope with the long prediction chain effectively. Most existing NVC models struggle to address the temporal error accumulation problem therein. To alleviate this issue,  many of them \cite{lu2019dvc,lu2020end,lu2020content, hu2021fvc, hu2020improving} rely on using a small intra-period setting (e.g. 10 or 12) to insert high-quality intra-frame more frequently. However, a smaller intra-period setting harms the whole compression efficiency. For example,  \cite{sheng2021temporal, li2022hybrid} show that intra-period
32 has an average of 23.8\% bitrate saving over intra-period 12 for H.265/HM. Thus, traditional standard committee \cite{bossen2013common} strictly defines the intra-period as --1, namely only single intra-frame is encoded for the whole video.
We think NVC should also follow this setting. This also makes it more fair when comparing NVC and traditional codec. 
However, we find previous SOTA DCVC-DC has large quality degradation under the intra-period --1 setting, as shown in Fig. \ref{first_page}. To solve this problem, two countermeasures are proposed.  One is increasing the video frame number to better learn the long-distance temporal correlation during the training. % But straightforwardly increasing frame number will cause the training very unstable, so we carefully adjust the gradient update strategy to stabilize the training.
  Another is that we propose modulating the temporal feature by periodically refreshing it, which can significantly alleviate the error propagation.  

With these effective Feature Modulation techniques, we build a new codec DCVC-FM, based on DCVC-DC.
In addition, our DCVC-FM involves other improvements toward a versatile NVC. Most existing NVCs are only optimized for RGB colorspace. Actually, traditional codecs and practical applications mainly adopt YUV colorspace. To this end, we design an NVC that can support both RGB and YUV without any fine-tuned training.  In addition, via improved implementation, this paper also showcases the low-precision inference which can significantly reduce the running time and memory cost with a negligible compression ratio degradation. Experiments show that our DCVC-FM can outperform VTM by 25.5\% under intra-period --1 setting, and also achieve non-trivial advantage over ECM. When compared with previous SOTA NVC DCVC-DC, 29.7\% bitrate saving is achieved while the MACs
(multiply–accumulate operations) are reduced by 16\%.

 In summary, our contributions are:

\begin{itemize}
    \item We modulate the latent feature via learnable quantization scaler, where a uniform quantization parameter sampling mechanism is proposed to help its learning. It enables our DCVC-FM to support a wide quality range in a single model, and the rate control capability is demonstrated. 
    \item  We not only exploit the training with longer video but also  module the temporal feature with a periodically refreshing mechanism to boost the quality. These help our DCVC-FM to tackle the long prediction chain.
    \item To further improve the practicality, we enable DCVC-FM to support both RGB and YUV colorspaces within a single model. Moreover, we demonstrate the low-precision inference with negligible bitrate increase.
    \item Our DCVC-FM can outperform all traditional codecs under intra-period --1 setting. When compared with the previous SOTA NVC, 29.7\% bitrate reduction is achieved with 16\% MAC reduction. Our codec is an important milestone in the development of NVC.
\end{itemize}

\section{Related Work}
\label{sec:related_work}

\subsection{Neural Image Compression}
Most recent neural image codec (NIC)  models follow hyperprior \cite{balle2018variational} and adopt a hierarchical framework design. %, where the parameters of entropy coding are first learned and transmitted.
 Some works \cite{qian2022entroformer,koyuncu2022contextformer, zou2022devil, liu2023learned} use transformer to strengthen the autoencoder or entropy model. Recently, the diffusion model \cite{yang2022lossy,goose2023neural,theis2022lossy,pan2022extreme} is also explored to improve the generation ability. In addition, the light-weight models \cite{wang2023evc,yang2023computationally} are also proposed. Now, NIC is quite powerful and its standardization process is already under consideration \cite{ascenso2023jpeg}.

\subsection{Neural Video Compression}
The success of NIC also pushes the development of NVC. The early  DVC \cite{lu2019dvc} follows a traditional residual coding-based framework, and uses NIC  to code the motion vector and residual, separately. Many NVCs \cite{hu2022coarse,Rippel_2021_ICCV,lu2020end,lin2020m,agustsson2020scale, Djelouah_2019_ICCV,liu2020neural, liu2023mmvc, ma2024uncertainty} also adopt this paradigm and design stronger sub-modules to boost the compression ratio. For example, the optical flow estimation in scale space \cite{agustsson2020scale} is proposed to handle the complex motion area.
Multiple reference frames are utilized to improve the temporal prediction \cite{lin2020m}. %Hu \textit{et al.} \cite{hu2022coarse} designed hyperprior-guided mode prediction. 
The block-based prediction mode selection is proposed in \cite{liu2023mmvc}.

When compared with residual coding, the conditional coding \cite{liu2020conditional, ladune2021conditional,canfvc,vct, li2021deep,sheng2021temporal, li2022hybrid, qi2023motion, li2023neural} shows larger potential because its temporal context is not limited to the predicted frame in pixel-domain and it does not rely on the sub-optimal subtraction to reduce redundancy. The feature-domain temporal context can be flexibly designed and its correlation with the current frame can be automatically learned.  In  \cite{liu2020conditional}, temporally conditional entropy models is designed. The DCVC series \cite{li2021deep,sheng2021temporal, li2022hybrid, qi2023motion, li2023neural}
propose using high-dimension context to improve encoding, decoding, as well as the entropy modelling.
The DCVC-TCM \cite{sheng2021temporal} introduces temporal feature propagation.
The DCVC-HEM \cite{li2022hybrid} designs a powerful entropy model utilizing both spatial and temporal contexts. The latest DCVC-DC \cite{li2023neural} already outperforms the under-developing ECM  by continually boosting the context diversity.
 
However,  there still exist several critical problems that block the practicality of NVC. The first is the quality range problem. Although DCVC-HEM and DCVC-DC support variable bitrates in a single mode, their quality range is quite limited and cannot meet the various quality requirements. In addition, most existing NVCs including previous SOTA DCVC-DC  still use a small intra-period setting (e.g., 10, 12, and 32), which is far from a practical scenario.  By contrast, our proposed DCVC-FM solve both of these two critical problems via feature modulation. 

\section{Proposed Method}
\subsection{Overview}

\begin{figure}[t]
		\begin{center}
			\includegraphics[width=1.08\linewidth]{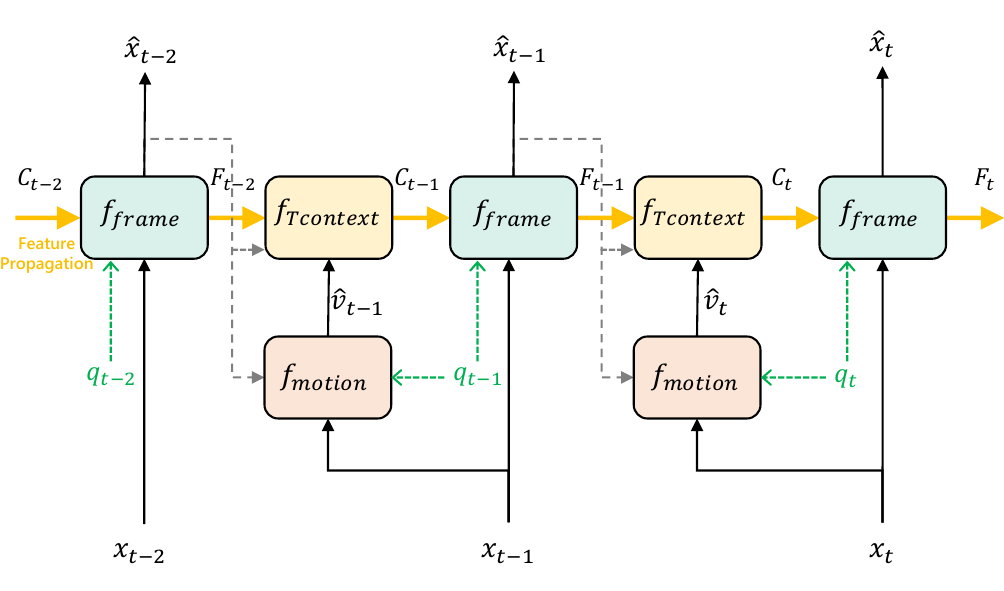}
		\end{center}
		\vspace{-0.4cm}
		\caption{The  framework of our DCVC-FM, built on DCVC-DC.	}
		\vspace{-4mm}
		\label{all_framwork}
	\end{figure}

Our DCVC-FM adopts the conditional coding-based framework and is built on DCVC-DC \cite{li2023neural}.  The overall framework is shown in Fig. \ref{all_framwork}.  For coding each input frame $x_t$ ($t$ is the frame index), there are three primary functions:  $f_{motion}$, $f_{Tcontext}$, and $f_{frame}$. At first,  $f_{motion}$ employs an optical flow network to estimate the motion vector $v_t$ between $x_t$ and the previous reconstructed frame $\hat{x}_{t-1}$. $v_t$ needs to be encoded, transmitted, and decoded as   $\hat{v}_{t}$. 
Subsequently, $f_{Tcontext}$ uses $\hat{v}_{t}$ to extract the temporal context $C_{t}$ from the propagated feature $F_{t-1}$ coming from the previous frame. 
Finally, conditioned on the motion-aligned temporal context $C_{t}$, $x_t$ is encoded, transmitted, and decoded as $\hat{x}_{t}$ via the function $f_{frame}$. Meanwhile, $f_{frame}$ generates the $F_{t}$ for the need of the next frame. When compared with  DCVC-DC, we make the supporting quality range increase from 3.8 dB to 11.4 dB (Section \ref{sec_quality_range}). %For $f_{motion}$ and $f_{frame}$, the quantization parameter $q_t$ can seamlessly adjust the quality level. 
In addition, to effectively cope with the intra-period --1 setting, we improve the feature propagation mechanism (Section \ref{sec_long_prediction}). Section \ref{sec_impl} shows the details of new capabilities: the single model for both RGB and YUV colorspaces, and the low-precision inference via our improved implementation.

\subsection{Wide Quality Range in a Single Model}
\label{sec_quality_range}

\begin{figure}[t]
		\begin{center}
			\includegraphics[width=0.95\linewidth]{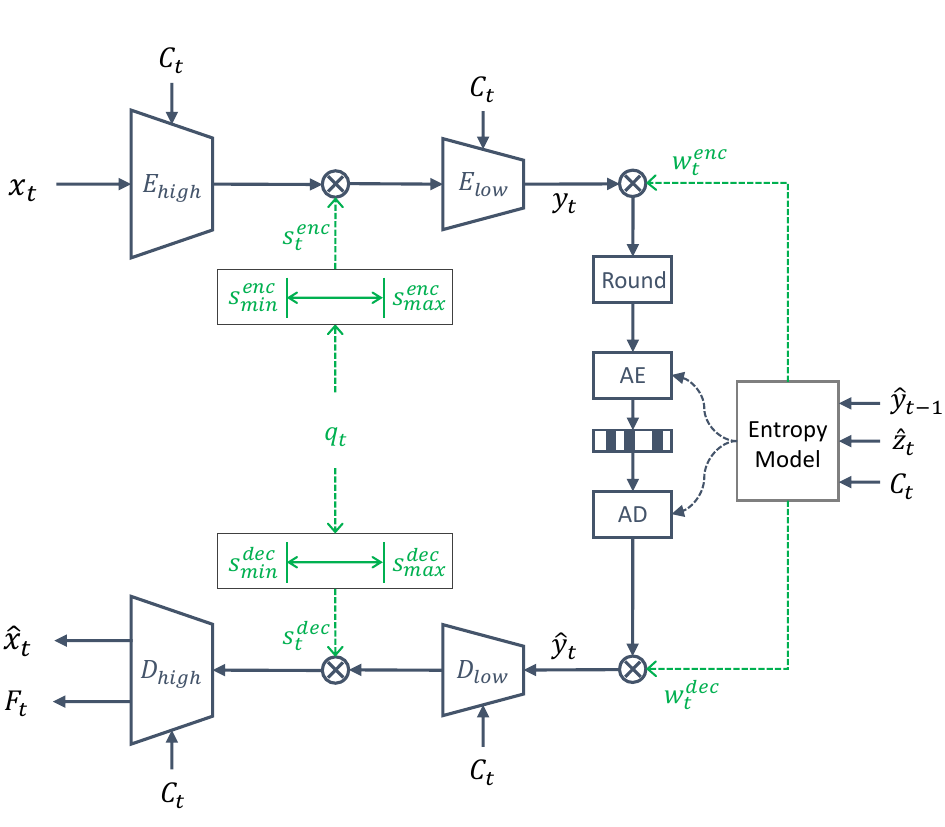}
		\end{center}
		\vspace{-0.3cm}
		\caption{The framework of our frame coding function $f_{frame}$. $E_{high}$   and $E_{low}$  are encoder  at high and low resolution, respectively. $D_{high}$ and $D_{low}$ are corresponding decoder. AE and AD are arithmetic encoder and decoder.	The quantization and inverse quantization processes are also applied to $f_{motion}$ in a similar way.}
		%\vspace{-5mm}
		\label{enc_dec}
	\end{figure}
 
%For both traditional and neural codecs, the optimization object is usually $Loss=\lambda \cdot D+ R$ \footnote{Sometimes $\lambda$ is applied to $R$.}, where $D$ and $R$ represent the distortion and bitrate, respectively. The $\lambda$ is used to control the trade-off between $D$ and $R$. At present, most existing NVCs need to train separate model for each quality target, and only provide a few (typically 4) different quality supports. This severely damages the practicality. Later, DCVC-HEM and DCVC-DC introduce multi-granularity quantization to enable variable quality levels in a single model, but the supporting quality range is only about 3 dB. So we 
A basic quantization and inverse quantization processes can be formulated as:
\begin{equation}
\label{eq_quant}
\small
\hat{I} = QS \cdot \lfloor \frac{I}{QS}\rceil
\end{equation}
$I$ and $QS$ are input value and quantization step, respectively. $\lfloor \cdot \rceil$ is the rounding operation.  The $QS$ controls the reconstruction quality of output $\hat{I}$. To enable variable quality for NVC, we also incorporate similar mechanism into the encoding and decoding processes to modulate the latent feature. The key challenge is how to decide the corresponding $QS$ and also let it support a wide range. 

Fig. \ref{enc_dec} shows our design for the frame coding function $f_{frame}$. As shown in this figure, during the encoding process, there are two values related to the quantization step. One is $s_{t}^{enc}$ and the other is $w_{t}^{enc}$. Both of them are used to modulate the latent feature, but they are generated in different ways and applied in different granularities.  $s_{t}^{enc}$ is the global quantization scaler and is generated according to the quantization parameter $q_t$ from the user input, which is similar with the concept of QP in the traditional codec. The $q_t$ is an integer scalar ranging from [0, $q\_num$--1], where  $q\_num$ is the adjustable number of   $q_t$ value and is set as 64 in the implementation.   
Similar to the traditional video codec where the quantization step size exponentially grows with the linear increase in QP, we use the following equation to interpolate $s_{t}^{enc}$ based on $q_t$ and the quantization scale range [$s_{min}^{enc}$, $s_{max}^{enc}$] via:
\begin{equation}
     s_{t}^{enc} = s_{min}^{enc} \cdot (\frac{s_{max}^{enc}}{s_{min}^{enc}}) ^ {\frac{q_t}{q\_num-1}}
\end{equation}
To improve the numerical stability, we change it to:
\begin{equation}
\label{eq_qs}
     s_{t}^{enc} = e^{\ln{s_{min}^{enc}}+\frac{q_t}{q\_num-1}\cdot(\ln{s_{max}^{enc}}-\ln{s_{min}^{enc}})}
\end{equation}
In our design, $s_{min}^{enc}$ and $s_{max}^{enc}$ are learned during  training. % and other the other quantization scales are interpolated based on $q_t$ during training and inference. 

We know the optimization object of a codec is usually $Loss_{RD}=R + \lambda \cdot D$ \footnote{Sometimes $\lambda$ is applied to $R$.}, where $R$ and $D$ represent the bitrate and distortion, respectively. The $\lambda$ is used to control the trade-off between $R$ and $D$. 
At present, most existing NVCs need to train a separate model for each quality level with the corresponding predefined and constant $\lambda$ value. As for our model, to support variable quality levels, the $\lambda$ is also variable during the training. We predefine a $\lambda$ range as [$\lambda_{min}$, $\lambda_{max}$]. In our implementation,  [$\lambda_{min}$, $\lambda_{max}$] is set as [1, 768].  For each training step, we randomly select $q_t$ and interpolate the $\lambda$ value from this range. The interpolation is similar with Eq. \ref{eq_qs} and calculated as:
 \begin{equation}
\label{eq_lambda}
 \lambda = e^{\ln{\lambda_{min}}+\frac{q_t}{q\_num-1}\cdot(\ln{\lambda_{max}}-\ln{\lambda_{min}})}
\end{equation}

For each step, we uniformly sample the integral $q_t$ value from the range [0, $q\_num$-1], and then obtain the $\lambda$ value via Eq. \ref{eq_lambda}   to calculate the $Loss_{RD}$. If we compare Eq. \ref{eq_lambda} and Eq. \ref{eq_qs}, we can know that the [$\lambda_{min}$, $\lambda_{max}$] will guide the learning of [$s_{min}^{enc}$, $s_{max}^{enc}$] via the back-propagation of $Loss_{RD}$. By controlling the $\lambda$ value range, we can easily adjust  the value range of quantization scaler.  
By employing the uniformly sampling mechanism for  $q_t$, the codec is able to experience different $s_{t}^{enc}$ values and  explore various trade-offs between the  $R$  and $D$ during training. This enhances the harmonization between the encoding and quantization processes, resulting in that our codec is able to learn finely grained and controllable quantization scaler $s_{t}^{enc}$ to modulate the latent feature.

It should be noted that, not like  Eq. \ref{eq_quant}, our codec does not have the restriction that the same quantization step value must be used during encoding and decoding because our quantization and inverse quantization are performed at latent feature domain. Removing this restriction enables a larger flexibility to smoothly adjust the quality. Thus, the corresponding $s_{t}^{dec}$ are separately learned during the decoding. In addition, just because of removing this restriction, we can let $s_{t}^{enc}$ be  multiplied to modulate the latent feature rather than being a divisor during the encoding. This helps stabilize the training process by circumventing the potential division by zero issues.

However, $s_{t}^{enc}$ is the same for all spatial positions, which may ignore the different spatial characteristics for the video content. So we follow \cite{li2022hybrid, Huang_2022_CVPR} and use entropy model to learn spatial-channel-wise quantization scaler $w_{t}^{enc}$. $w_{t}^{enc}$ not only helps achieve precise modulation at each position but also is adaptive to the video content of each frame. This kind of content-adaptive and dynamic feature modulation can also improve the final compression efficiency.

\begin{figure}[t]
	\centering
	\includegraphics[width=1.0\linewidth]{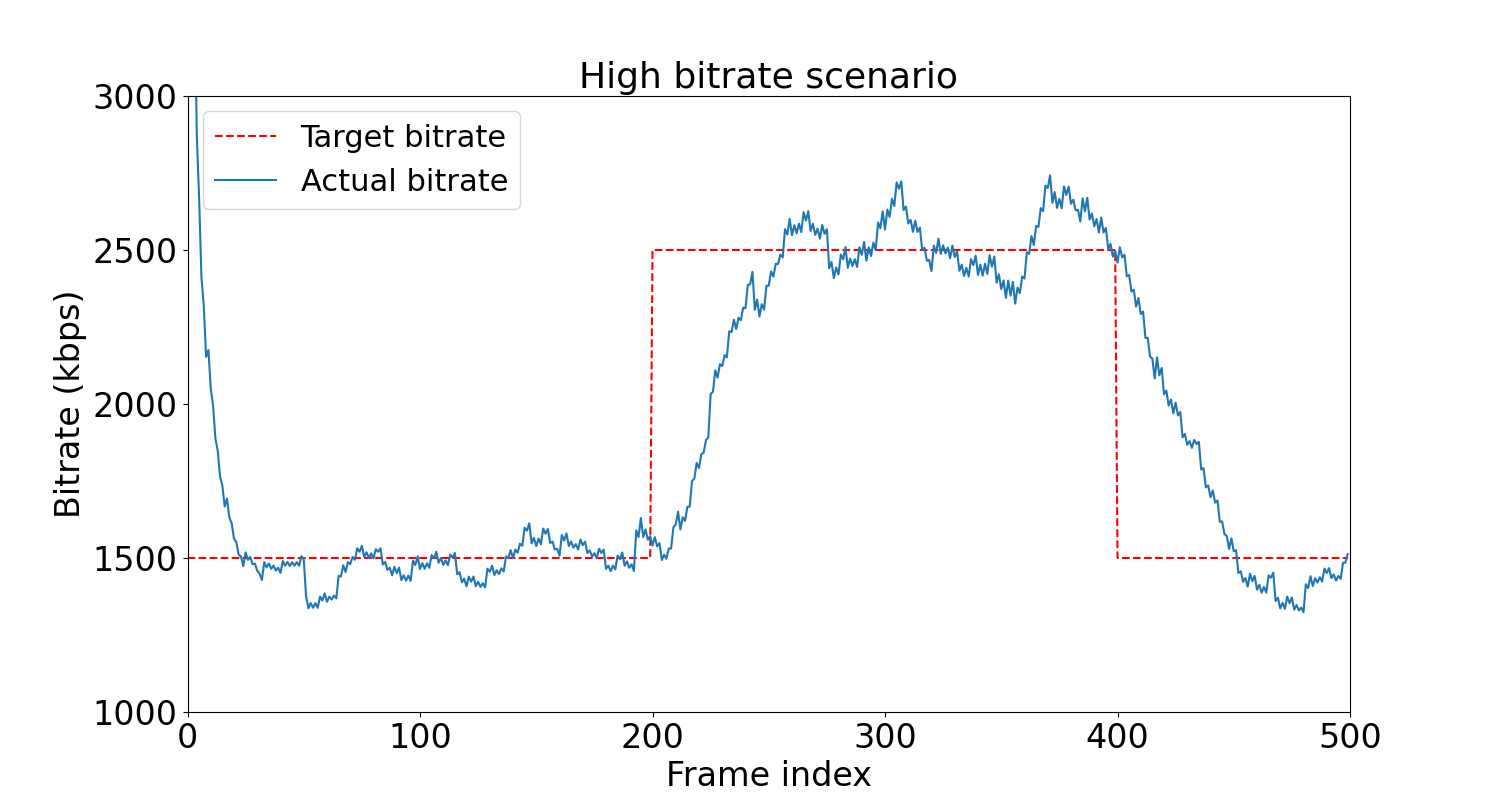} \\
 	\vspace{-0.1cm}
	\includegraphics[width=1.0\linewidth]{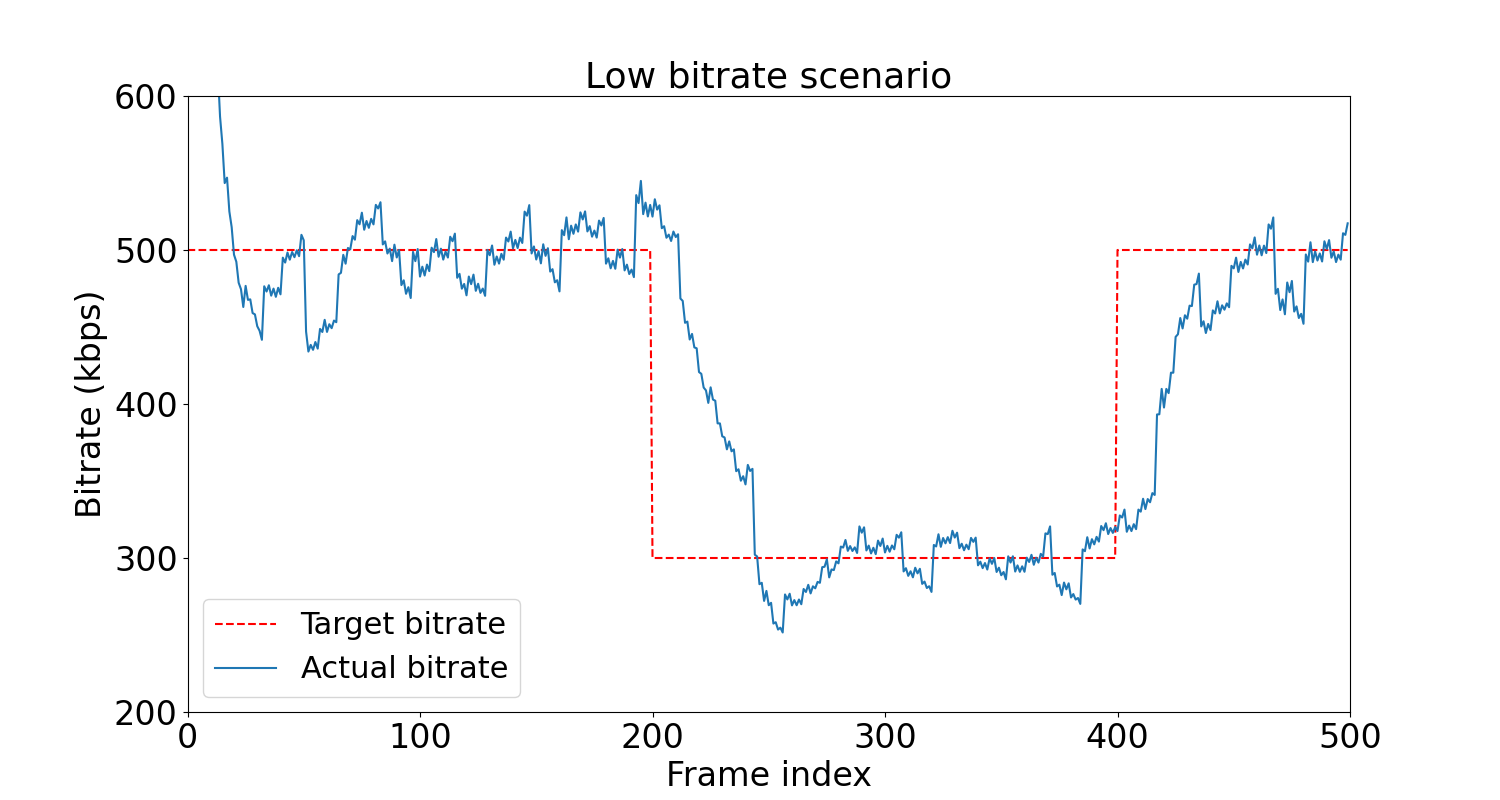}
	\vspace{-0.3cm}
	\caption{Rate control examples using \textit{BasketballDrive} video sequence (1080p, 50fps, 500 frames). The above (below) example is with a relatively high (low) target bitrate scenario. %The bitrate is smoothed with the previous frames in 1 second.
 }
		\vspace{-0.4cm}
	\label{fig_rc}
\end{figure}

Benefited from these advanced designs, our DCVC-FM finally can support quality adjustment in a wide range. This capability is also the prerequisite of rate control, which is the core functionality of a practical codec. Fig. \ref{fig_rc} shows two rate control examples with different target bitrate scenarios. As we can adjust the $q_t$ for each frame in our codec, the fluctuated target bitrate can be supported, as shown in Fig. \ref{fig_rc}. The actual bitrate gets close to the target bitrate. It is noted that we just showcase the feasibility of rate control using DCVC-FM rather than focusing on  designing new rate control algorithm for NVC. Currently, we just simply adjust the  $q_t$ based on the buffer fullness (more details are in supplementary materials). In the future, more advanced rate control algorithms can be proposed based on DCVC-FM.

\subsection{Long Prediction Chain}
\label{sec_long_prediction}

\begin{figure}[t]
		\begin{center}
			\includegraphics[width=1\linewidth]{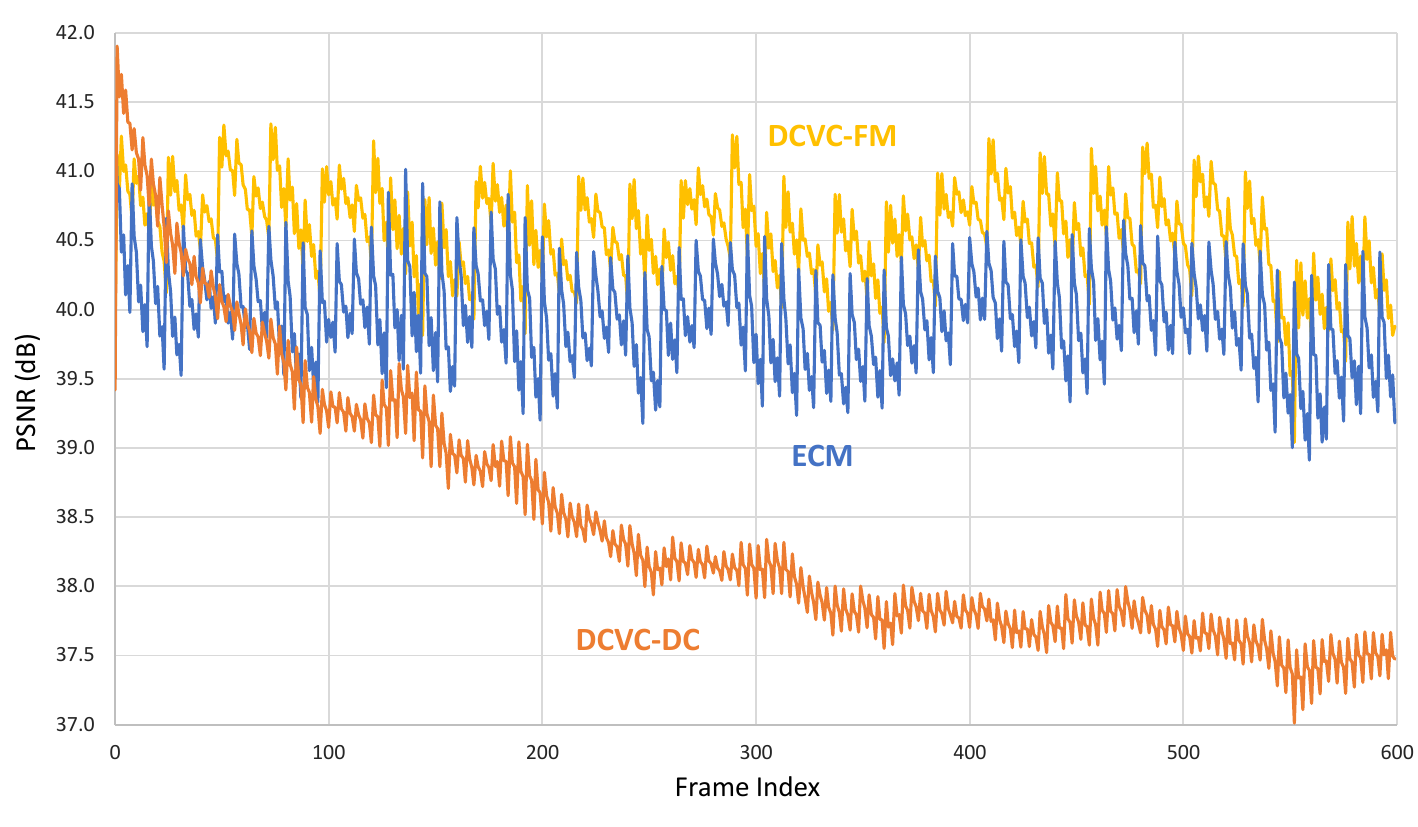}
		\end{center}
		\vspace{-0.3cm}
		\caption{ Quality comparison across frames.	The test video is \textit{KristenAndSara} from HEVC E dataset (video conferencing scenario). The average bpp (bits per pixel) results of DCVC-DC, ECM, and proposed DCVC-FM are 0.0037, 0.0029, and 0.0026, respectively. Intra-period = --1.}
		\vspace{-5mm}
		\label{fig_psnr}
	\end{figure}

The temporal quality degradation is a fundamental problem for all video codecs, but is especially serious for NVC. Previous NVCs already began to address this problem. For example, DCVC-DC follows traditional codec and  introduces the widely-used hierarchical quality structure to periodically improve the quality. This can help alleviate the error propagation but is not enough. As shown in Fig. \ref{fig_psnr}, DCVC-DC has serious quality degradation when only single intra frame is used (i.e. intra-period = --1). 
To solve this problem, two countermeasures are proposed. It should be noted that our NVC is conditional coding-based framework, whose major advantage is using temporal feature as the context and condition. In particular, the temporal feature can be propagated across many frames. So the key is how to design a more effective temporal feature propagation mechanisms. 
%for the example video from video conferencing scenario.  
%For such kind of video, how to effectively solve the temporal quality degradation problem is especially important .has relative small motion. 
Our first improvement is increasing the video frame number during the training. Although it is a simple modification, it is quite helpful. A longer video  enables recognizing similar patterns over long distances of time and then better exploring   the temporal correlation.
%The second improvement is periodically refreshing the propagated temporal feature. We know most existing NVCs rely on frequently inserting high-quality intra frames to cut off the accumulated error. As aforementioned, the price is significantly increasing the overall bitrate as these intra frames have large bitrate cost. Our workaround is frequently inserting high-quality inter frames with acceptable bitrate increase on these frames. 
\begin{figure}[t]
		\begin{center}
			\includegraphics[width=1\linewidth]{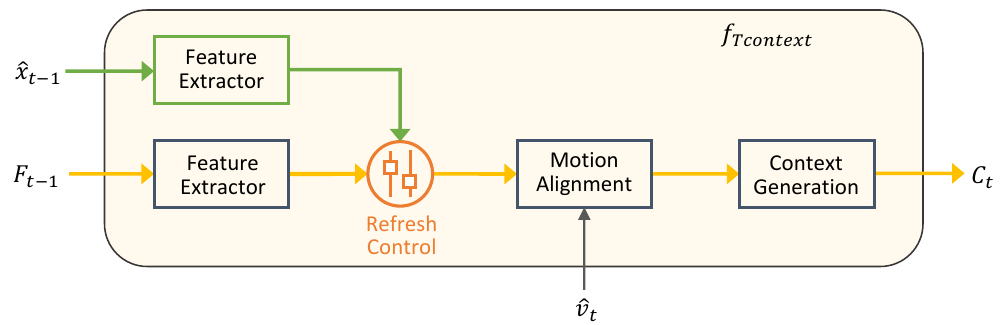}
		\end{center}
		\vspace{-0.4cm}
		\caption{ Temporal feature modulation via periodical refresh. }
		\vspace{-3mm}
		\label{fig_feature_refresh}
	\end{figure}
 
Feature propagation is  also double-edged sword, as the propagated feature may be contaminated by accumulated errors or contain some uncorrelated information.
To this end, we propose modulating the  the propagated feature, and force NVC refresh it periodically (the refresh period is set as 32 in the implementation). As shown in Fig. \ref{fig_feature_refresh}, for the input frame $x_t$, we will not let $f_{Tcontext}$ extract temporal context from the propagated feature  $F_{t-1}$ if we expect to perform the feature refresh. Instead, we use a separate feature extractor module to extract the  temporal context from $\hat{x}_{t-1}$. As $\hat{x}_{t-1}$ only has pixel information with 3 dimensions, which contains much less information than $F_{t-1}$. So, to facilitate the coding of $x_t$, it will force this separate module to extract correlated temporal context from $\hat{x}_{t-1}$ as much as possible. The new extracted temporal context will be  propagated to future frames. Such refresh-based modulation mechanism effectively alleviates the error propagation problem.  As shown in Fig. \ref{fig_psnr}, our DCVC-FM can maintain the quality across frames with lower bitrate cost when compared with DCVC-DC.

\subsection{Implementation}
\label{sec_impl}

\quad\,\textbf{Single model for both RGB and YUV colorspaces.} Although DCVC-DC supports a single network structure for both RGB and YUV colorspaces, the separate model weights with separate training are still required. To improve the versatility of NVC, we directly train a single model for both RGB and YUV without additional fine-tuning. To support it, our training loss covers both colorspaces as: $Loss_{RD}=R + \lambda \cdot (k \cdot D_{YUV} + (1-k) \cdot D_{RGB})$. $D_{YUV}$ and $D_{RGB}$ are the distortion in YUV and RGB, respectively. $k$ is the hyper-parameter for weighting, and set as 0.8 in the implementation. To use the same input interface for RGB and YUV,  the UV contents will be up-sampled if the input is  YUV420 content. Correspondingly, UV contents are
down-sampled after obtaining the reconstructed frame.

\textbf{Low precision inference.}  Most NVCs only report the results based on 32-bit floating point implementation. 
We hope to use 16-bit to accelerate the NVC.
However, to support 16-bit, we need to improve the implementation of \textit{grid\_sample} function, which is widely used for motion alignment. Our DCVC-FM is implemented with PyTorch.  
In the PyTorch \footnote{Currently up to PyTorch-2.1 version.}, the ``grid'' parameter in \textit{grid\_sample} represents the absolute position in the frame. When using 16-bit precision, only 10-bit significand is not enough to represent \footnote{https://en.wikipedia.org/wiki/Half-precision\_floating-point\_format}. %For example, the width of 1080p video is 1920 and 16-bit floating precision can only represent integer precision for position [1024, 2048], which means motion vector is also limited to integer precision.
To solve this problem, we use  16-bit precision  to represent the relative offset and reimplement the  \textit{grid\_sample}. %  \footnote{The reimplemented \textit{grid\_sample} codes are in supplementary materials.}.
 It enables the NVC inference at 16-bit precision, and can significantly save the memory and complexity with negligible compression ratio change.

%the warping process with relative offsets as input using CUDA could make the neural network run using 16-bit floating points without obvious compression ratio degradation.
\textbf{Structure optimization.} To reduce computation, we further make two additional improvements based on DCVC-DC. One is  reducing the convolution kernel size at the high-resolution feature for the motion estimation module. Another is that we use more depthwise separable convolutions which can  reduce the computation cost and  alleviate the over-fitting simultaneously \cite{chollet2017xception}. The section \ref{section_aba} verifies that these two improvements bring non-trivial MAC reduction with quite small bitrate increase.

\section{Experimental Results}
\subsection{Experimental Settings}

\quad\,\,\textbf{Datasets.} In line with most existing NVCs, we utilize the Vimeo-90k dataset \cite{xue2019video} for the training. Besides the ready-made 7-frame videos in Vimeo-90k, we also process the raw Vimeo videos \cite{Vimeo_link} to generate additional 6,658 videos, and each consists of 32 frames. The 7-frame video is used for training first, followed by the use of the 32-frame for fine-tuning, which further boosts the quality. For testing, we use the common HEVC B$\sim$E \cite{bossen2013common}, UVG \cite{mercat2020uvg}, and MCL-JCV \cite{wang2016mcl} datasets.

%\textbf{Model Training.} We extend the multi-stage training strategy proposed in \cite{li2023neural} with additional steps training using 32 frames sequences. 

%Different $\lambda$ values are used at different steps to support variable bitrate in single model. To support wider bitrate range, We set the minimum $\lambda$ value as 1 and maximum value as 768. The $\lambda$ values are randomly selected in between during training. Others are the same as \cite{li2023neural} and more details could be found in the supplementary materials.

\begin{table*}[t]
  \centering
  \caption{BD-Rate (\%) comparison in RGB colorspace. 96 frames with intra-period=32.}
  \centering
   \renewcommand{\arraystretch}{1.2}
    \small
    \begin{tabular}{ccccccccc}
    \toprule[1.0pt]
                                         & UVG    & MCL-JCV  & HEVC B & HEVC C & HEVC D & HEVC E      & Average    \\ \hline
VTM-17.0                        & 0.0      & 0.0    & 0.0       & 0.0    &  0.0    & 0.0     & 0.0     \\ \hline
HM-16.25                         & 38.4	    & 45.6	 & 40.3	     & 37.9	  &  32.4	& 41.0	  & 39.3    \\ \hline
ECM-5.0                        & --10.6	& --13.2 & --11.5    & --12.6 &  --11.2	& --9.8   & --11.5   \\ \hline
CANF-VC	 	                 & 73.0	    & 70.8	 & 64.3	     & 76.2	  &  63.1	& 120.5	  & 78.0    \\ \hline
DCVC  		             & 166.1    & 121.5	 & 123.0	 & 143.0  &  98.2	& 272.9	  & 154.1   \\ \hline
DCVC-TCM  		 & 44.1	    & 51.0	 & 40.2	     & 66.3	  &  37.0	& 82.7	  & 53.6    \\ \hline
DCVC-HEM  		     & 1.1	    & 8.6	 & 5.1	     & 22.2	  &  2.4	& 20.5	  & 10.0    \\ \hline
DCVC-DC	 	             & --19.1	& --11.3 & --12.0    & --10.3 &  --26.1	& --18.0  & --16.1  \\ \hline
DCVC-FM                                 & --17.0   & --5.6  & --14.3    & --23.7 &  --36.7	& --24.5  & --20.3  \\
    \bottomrule[1.0pt]
    \end{tabular}
    \vspace{0.1cm}
    \footnotesize{ \\  Note:  Some numbers are slightly different with those in \cite{li2023neural} as we are using a wider quality range to calculate BD-Rate.}
    \vspace{-0.0cm}
  \label{tab_rgb_psnr}
\end{table*}

\begin{table*}[t]
  \centering
  \caption{BD-Rate (\%) comparison in YUV420 colorspace. 96 frames with intra-period=32.  }
   \renewcommand{\arraystretch}{1.2}
    \small
    \begin{tabular}{ccccccccc}
    \toprule[1.0pt]
                                         & UVG    & MCL-JCV  & HEVC B & HEVC C & HEVC D   & HEVC E        & Average        \\ \hline

						  VTM-17.0	    & 0.0      & 0.0    & 0.0       & 0.0    &  0.0     & 0.0       & 0.0            \\ \hline
						  HM-16.25	    &38.0	   &45.9	&39.3	    &34.6	 &  29.0	&37.5		&37.4            \\ \hline
						  ECM-5.0	    &--11.5	   &--15.0	&--12.7	    &--13.7	 &--12.2	&--11.0		&--12.7          \\ \hline
						  DCVC-DC       &--17.8	   &--12.0	&--10.8	    &--12.4	 &--28.5	&--20.4		&--17.0          \\ \hline
						  DCVC-FM      &--21.6	   &--11.4	&--16.3	    &--25.8	 &--39.3	&--30.1		&--24.1          \\

    \bottomrule[1.0pt]
    \end{tabular}
    \vspace{-0.0cm}
  \label{tab_yuv_psnr_96f_ip32}
\end{table*}

\textbf{Test Conditions.} All tests are conducted under low-delay coding settings. We employ the BD-Rate metric \cite{bjontegaard2001calculation} to measure the compression ratio change, where positive values are bitrate increase, and negative values indicate bitrate savings. The video quality is evaluated using PSNR. Our benchmarks include traditional codecs H.265/HM \cite{HM}, H.266/VTM \cite{VTM}, and ECM \cite{ECM}.

For the RGB colorspace, to facilitate the comparison with existing methods, we follow the test condition in \cite{li2023neural} and test 96 frames for each video with intra-period 32. We also compare our DCVC-FM with previous SOTA NVC models, including CANF-VC \cite{canfvc}, DCVC \cite{li2021deep}, DCVC-TCM \cite{sheng2021temporal}, DCVC-HEM \cite{li2022hybrid}, and DCVC-DC \cite{li2023neural}.

For the YUV420 colorspace, we initially follow \cite{li2023neural} and test 96 frames with  intra-period 32. As aforementioned, such a small intra-period is far from practical applications, and the traditional standard committee \cite{bossen2013common} strictly sets the intra-period as --1. Thus, we advocate for testing NVCs under this intra-period --1 setting. Unlike \cite{li2023neural}, which only tests 96 frames, we also test all frames with intra-period --1.  It is quite challenging for existing NVCs but is the most fair setting for the comparison with traditional codec.
It is noted that all coding tools and reference structure of traditional codecs use the best settings to represent their best compression ratio. Both ECM-5.0 and the recent ECM-11.0 are tested.
For NVC, most existing models are only optimized for the RGB colorspace, and DCVC-DC \cite{li2023neural} is the only NVC that has released a model for the YUV420 colorspace. The BD-Rate calculations are based on the weighted PSNR for the three color components, with weights of (6, 1, 1) / 8, consistent with the standard committee \cite{sullivan2011meeting}.

\begin{table*}[t]
  \centering
  \caption{BD-Rate (\%) comparison in YUV420 colorspace. 96 frames with intra-period = --1. }
   \renewcommand{\arraystretch}{1.2}
    \small
    \begin{tabular}{ccccccccc}
    \toprule[1.0pt]
                                         & UVG    & MCL-JCV  & HEVC B & HEVC C & HEVC D   & HEVC E        & Average        \\ \hline

						  VTM-17.0	    & 0.0      & 0.0    & 0.0       & 0.0    &  0.0     & 0.0       & 0.0            \\ \hline
						  HM-16.25	    &39.5	   &48.3	&41.5	    &40.3	 &  32.6	&41.5		&40.6            \\ \hline
						  ECM-5.0	    &--13.3	   &--16.4	&--14.6	    &--15.6	 &--14.1	&--12.6		&--14.4          \\ \hline
						  DCVC-DC       &--13.6	   &--7.9	&--7.8	    &--5.6	 &--27.6	&--10.3		&--12.2          \\ \hline
			%			  DCVC-DC (FR)  &--18.9	   &--12.2	&--10.4	    &--9.3	 &--30.0	&--18.2		&--16.5          \\ \hline
						  DCVC-FM      &--25.4	   &--11.6	&--17.1	    &--24.4	 &--41.5	&--31.6		&--25.3          \\

    \bottomrule[1.0pt]
    \end{tabular}
    \vspace{-0.0cm}
  \label{tab_yuv_psnr_96f}
\end{table*}

\begin{table*}[t]
  \centering
  \caption{\textbf{BD-Rate (\%) comparison in YUV420 colorspace. All frames with intra-period = --1.}   }
   \renewcommand{\arraystretch}{1.2}
    \small
    \begin{tabular}{ccccccccc}
    \toprule[1.0pt]
                                         & UVG    & MCL-JCV  & HEVC B & HEVC C & HEVC D   & HEVC E        & Average        \\ \hline

						  VTM-17.0	    & 0.0      & 0.0    & 0.0       & 0.0    &  0.0     & 0.0       & 0.0            \\ \hline
						  HM-16.25	    &40.1	   &48.6	&47.6	    &41.0	 &  34.5	&42.8		&42.4            \\ \hline
						  ECM-5.0	    &--14.9	   &--17.0	&--17.3	    &--16.6	 &--16.1	&--14.1		&--16.0          \\ \hline
                            ECM-11.0	  &--20.0	 &--22.1	&--22.2   &--21.2  &--20.4	  &--17.2	  &--20.5          \\ \hline
						  DCVC-DC       &5.7	   &--5.0	&12.2	    &--4.2   &--16.5	&84.4		&12.8            \\ \hline
				%		  DCVC-DC (FR)  &--15.8	   &--12.6	&--9.2	    &--19.6	 &--28.7	&--13.3		&--16.5          \\ \hline
						  \textbf{DCVC-FM}      &\textbf{--20.7}	   &\textbf{--10.3}	&\textbf{--18.2}	    &\textbf{--32.2}	 &\textbf{--41.2}	&\textbf{--30.2}		&\textbf{--25.5}          \\

    \bottomrule[1.0pt]
    \end{tabular}
    \vspace{-0.0cm}
  \label{tab_yuv_psnr_allf}
\end{table*}

\begin{figure*}[t]

\minipage{0.33\textwidth}
\includegraphics[width=1.07\linewidth,height=0.9\linewidth]{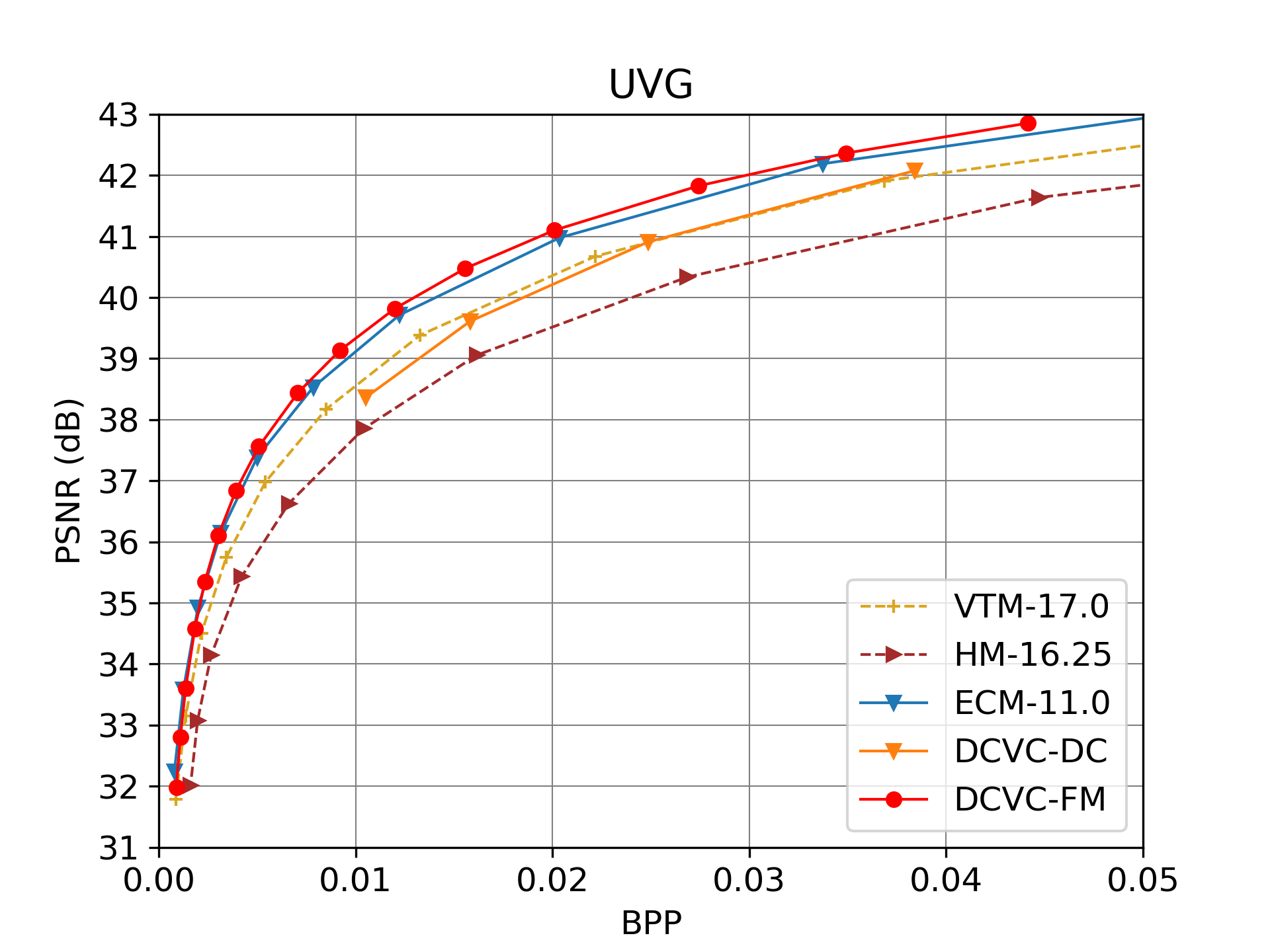}
\endminipage
\minipage{0.33\textwidth}
\includegraphics[width=1.07\linewidth,height=0.9\linewidth]{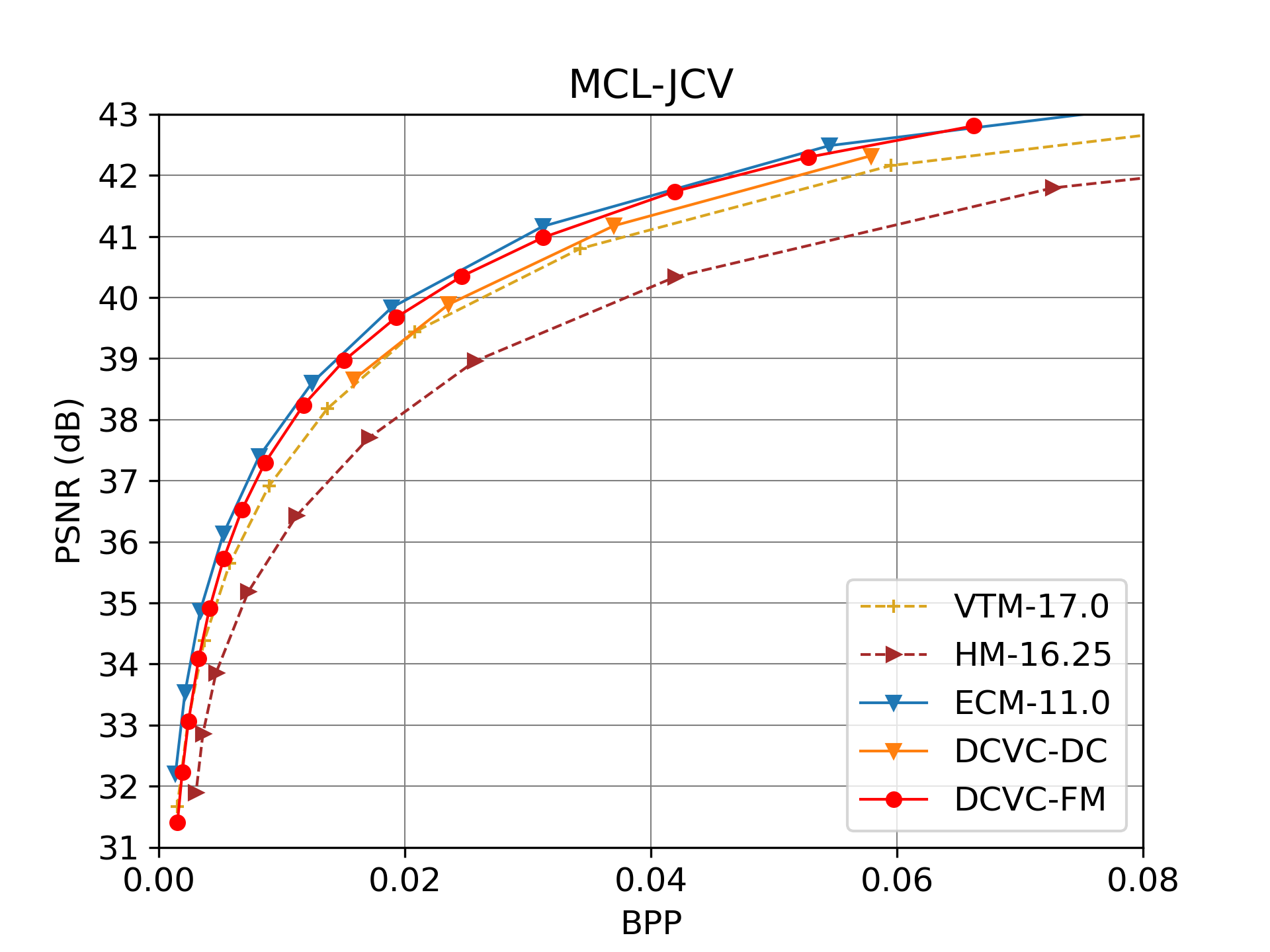}
\endminipage
\minipage{0.33\textwidth}%
\includegraphics[width=1.07\linewidth,height=0.9\linewidth]{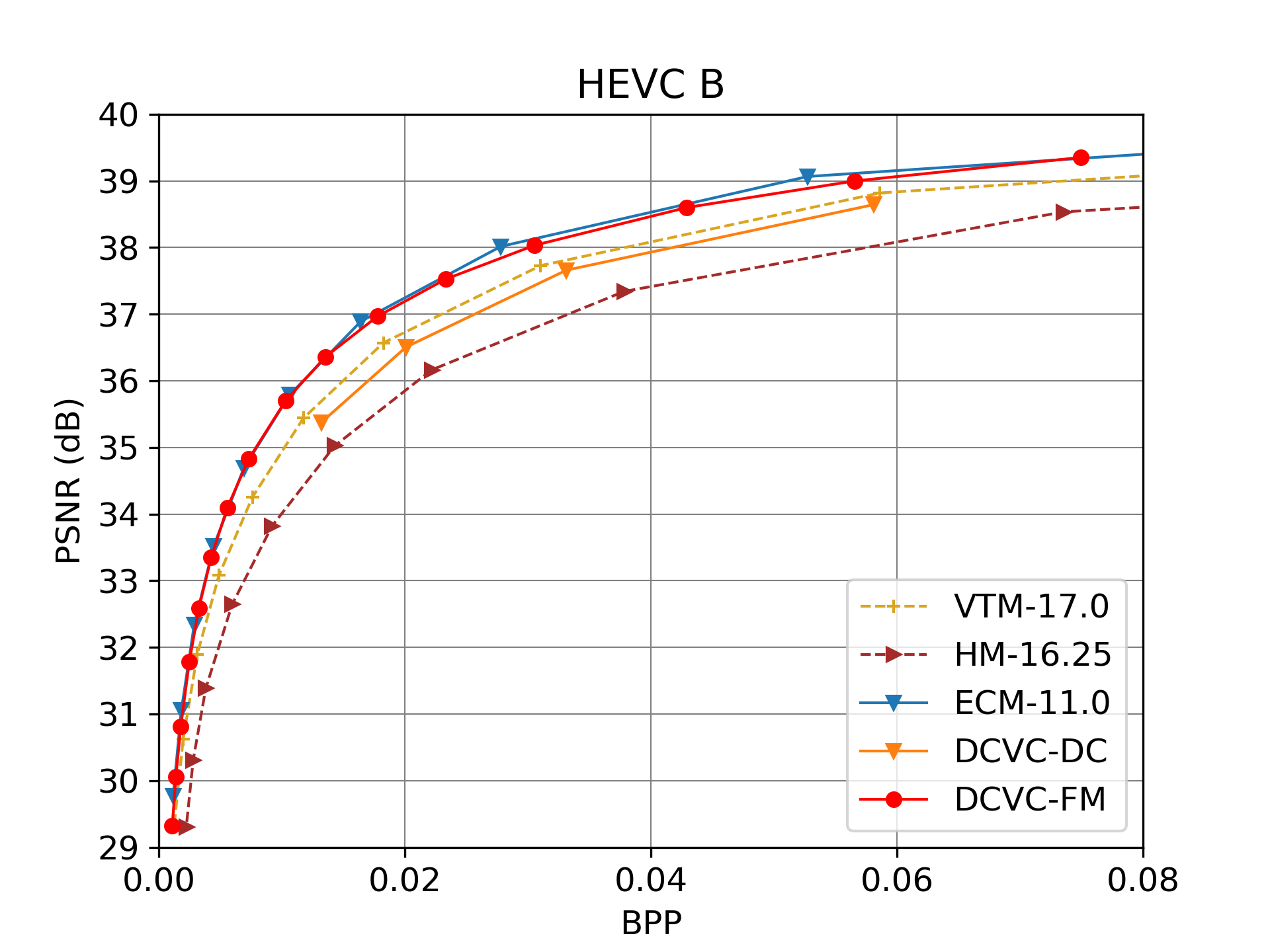}
\endminipage

% \minipage{0.33\textwidth}
%   \includegraphics[width=\linewidth]{figures/yuv420_96f_ip96_rd/yuv420_96f_ip96_MCL-JCV.png}
% \endminipage
% \minipage{0.33\textwidth}
%   \includegraphics[width=\linewidth]{figures/yuv420_96f_ip96_rd/yuv420_96f_ip96_MCL-JCV_low.png}
% \endminipage
% \minipage{0.33\textwidth}%
%   \includegraphics[width=\linewidth]{figures/yuv420_96f_ip96_rd/yuv420_96f_ip96_MCL-JCV_high.png}
% \endminipage

% \minipage{0.33\textwidth}
%   \includegraphics[width=\linewidth]{figures/yuv420_96f_ip96_rd/yuv420_96f_ip96_HEVC_B.png}
% \endminipage
% \minipage{0.33\textwidth}
%   \includegraphics[width=\linewidth]{figures/yuv420_96f_ip96_rd/yuv420_96f_ip96_HEVC_B_low.png}
% \endminipage
% \minipage{0.33\textwidth}%
%   \includegraphics[width=\linewidth]{figures/yuv420_96f_ip96_rd/yuv420_96f_ip96_HEVC_B_high.png}
% \endminipage
  \vspace{-0.0cm}
%, MCL-JCV, and HEVC B
\caption{Rate and distortion curves for UVG , MCL-JCV, and HEVC B datasets. The comparison is in YUV420 colorspace. All frames with intra-period = --1.
%Each row shows a dataset.
%From left to right the figures are overall quality range, low quality range and high quality range, respectively. 
The curves of more datasets are in supplementary materials.}
\label{fig_yuv_psnr_allf_curve}
\end{figure*}

\begin{table*}[t]
  \centering
  \caption{The quality range (PSNR, dB) comparison in YUV420 colorspace. All frames with intra-period = --1.   }
   \renewcommand{\arraystretch}{1.2}
    \small
    \begin{tabular}{ccccccccc}
    \toprule[1.0pt]
                                         & UVG    & MCL-JCV  & HEVC B & HEVC C & HEVC D   & HEVC E        & Average        \\ \hline

						  DCVC-DC       &3.7	   &3.7	     &3.3	    &4.2      &4.6	&3.6		&3.8            \\ \hline
						  DCVC-FM      &10.9	   &11.4	&10.0	    &12.4	 &12.5	&11.3		&11.4           \\

    \bottomrule[1.0pt]
    \end{tabular}
    \vspace{-0.3cm}
  \label{tab_quality_range_allf}
\end{table*}

\subsection{Comparisons with Previous SOTA Methods}

\quad\,\textbf{RGB colorspace.} Table \ref{tab_rgb_psnr} shows the performance comparison for RGB colorspace under 96 frames with intra-period 32. From this table, we can see that our DCVC-FM obtains the best compression ratio. Our DCVC-FM achieves an average of 20.3\% bitrate saving over VTM, and also has significant advantage over ECM. In addition, our DCVC-FM is also better DCVC-DC as the bitrate saving of DCVC-DC is 16.1\% over VTM.
Under this setting, still outperforming DCVC-DC is not an trivial thing considering that our codec has much wider quality range and support both RGB and YUV in single model when compared with DCVC-DC.

\textbf{YUV420 colorspace.} Table \ref{tab_yuv_psnr_96f_ip32} shows the comparison for YUV420 colorspace under 96 frames with intra-period 32. As shown in this table, our DCVC-FM can surpass all traditional codecs and DCVC-DC. Our bitrate saving over VTM is 24.1\% on average while that of DCVC-DC is 17.0\%. Table \ref{tab_rgb_psnr} and Table \ref{tab_yuv_psnr_96f_ip32} show the consistent improvement over DCVC-DC for both RGB and YUV420 colorspaces.

However, we focus more on the evaluation using intra-period --1 setting. Table \ref{tab_yuv_psnr_96f} shows the comparison under 96 frames with intra-period --1. From this table, we can find that our DCVC-FM  can maintain the bitrate saving, i.e. 25.3\%, over VTM, when compared with the number 24.1\% under intra-period 32. By contrast, the bitrate saving of DCVC-DC over VTM decreases from 17.0\% to 12.2\% when the intra-period changes from 32 to --1.

But only testing 96 frames is not enough to evaluate the behavior of NVC when handling the long prediction chain. So we also test the all frames for each test video. %, same with the test condition in traditional standard committee.
Table \ref{tab_yuv_psnr_allf} shows the corresponding comparison. From this table, we can see that our DCVC-FM still can outperform VTM by 25.5\%. By contrast, DCVC-DC has large performance drop, i.e. 12.8\% bitrate increase over VTM, when testing all frames. In particular, DCVC-DC has 84.8\% bitrate increase for HEVC E dataset. By contrast, our codec can cope with this dataset well. 
In addition, if we use DCVC-DC as anchor in Table \ref{tab_yuv_psnr_allf}, our DCVC-FM can achieve an average of 29.7\% bitrate saving for all datasets.
As far as we know, our DCVC-FM is the first NVC that can achieve such high compression ratio under intra-period --1 setting.

Fig. \ref{fig_yuv_psnr_allf_curve} shows the rate-distortion curves. From these curves, we can see that our DCVC-FM achieves much wider quality range than DCVC-DC. For example, the quality range of DCVC-DC on MCL-JCV dataset is 3.7 dB ([38.66, 42.32]), which cannot meet the requirement of higher compression ratio. By contrast,  DCVC-FM is 11.4 dB ([31.40, 42.80]), which is much wider. Table \ref{tab_quality_range_allf} shows more comparisons and we can see the obvious quality range expansion on each test dataset. The quality range of the proposed model is similar to that encoded with VTM using QP 25$\sim$49.

\subsection{Ablation Study}
\label{section_aba}

\begin{table}[t]
\caption{Ablation study using BD-Rate (\%). }
\centering
\scalebox{0.95}{
\renewcommand{\arraystretch}{1}
\begin{tabular}{ccccccc}
\toprule[1.0pt]
          
                                    &  { $M_a$}  & $M_b$                      & $M_c$                    & $M_d$                     & $M_e$                      & $M_f$                           \\ \hline
              Structure             &            &   \multirow{2}*{\checkmark}& \multirow{2}*{\checkmark}& \multirow{2}*{\checkmark} & \multirow{2}*{\checkmark}  &  \multirow{2}*{\checkmark}       \\     
          optimization              &            &                            &                          &                           &                            &                                  \\ \hline    
   Wider quality                    &            &                            & \multirow{2}*{\checkmark}& \multirow{2}*{\checkmark} & \multirow{2}*{\checkmark}  &  \multirow{2}*{\checkmark}       \\     
    range support                   &            &                            &                          &                           &                            &                                  \\ \hline    
   Single model                     &            &                            &                          & \multirow{2}*{\checkmark} & \multirow{2}*{\checkmark}  &  \multirow{2}*{\checkmark}       \\     
   for RGB\&YUV                     &            &                            &                          &                           &                            &                                  \\ \hline    
   Training with                    &            &                            &                          &                           & \multirow{2}*{\checkmark}  &  \multirow{2}*{\checkmark}       \\     
   longer video                     &            &                            &                          &                           &                            &                                  \\ \hline    
   Feature                          &            &                            &                          &                           &                            & \multirow{2}*{\checkmark}        \\     
     refresh                          &            &                            &                          &                           &                            &                                 \\ \hline          
{ BD-Rate(\%)}                       & 0.0        &         0.8                &        3.4               &       4.8               &       --15.5                &  --29.7                         \\     
\bottomrule[1.0pt] 
\end{tabular}
}
	\vspace{-3mm}
\label{tab_abalation}
\end{table}

Table \ref{tab_abalation} shows the ablation study on each improvement. In this table, the baseline model $M_a$ is DCVC-DC. We first test the structure optimization on reducing motion estimation module and using more depthwise convolution. This optimization ($M_b$) has 0.8\% bitrate increase but can save the MACs by 16\% reduction over DCVC-DC ($M_a$), as shown in Table \ref{tab_complexity}. Table \ref{tab_abalation} also shows that, when further enabling wider quality range support ($M_c$), there is 3.4\% bitrate increase. If also supporting single model for both RGB and YUV colorspaces ($M_d$), the bitrate increase gets to 4.8\% because these functionalities are not cost-free.     Training with longer video  can bring a large performance improvement, and $M_e$ can achieve 15.5\% bitrate saving, which demonstrates the benefits of utilizing longer temporal correlation. Based on $M_e$, refreshing the temporal feature ($M_f$) makes the bitrate saving increase to 29.7\% as it can effectively alleviate the quality degradation problem.

Table \ref{tab_complexity} also shows the runtime comparison. Our MACs have obvious reduction but the actual running time using 32-bit floating point inference is a little higher. This is because we use more depthwise convolution layers which have a lower computational density than the normal convolution, but can be further accelerated in the future \cite{lu2021optimizing}.   When enabling 16-bit floating point inference with our optimized implementation for \textit{grid\_sample}, the running time has significant reduction. In particular, our NVC can save half of the memory usage if using 16-bit inference.   Table \ref{tab_fp16_comp} also shows that there is a significant 87.3\% bitrate increase without our optimized implementation.

 \begin{table}[t]
  \centering
  \caption{BD-Rate (\%) using 16-bit floating point (fp) inference.}
   \renewcommand{\arraystretch}{1.2}
    \small
    \begin{tabular}{ccc }
    \toprule[1.0pt]
                    fp32        & fp16 w/o optimization  & fp16 w/ optimization       \\ \hline
				     0.0         & 87.3                    & 0.9            \\

    \bottomrule[1.0pt]
    \end{tabular}
    \vspace{-0.0cm}
  \label{tab_fp16_comp}
\end{table}

\begin{table}[t]
\caption{Complexity comparison.}
\centering
\scalebox{0.84}{
\renewcommand{\arraystretch}{1.2}
\begin{tabular}{ccccc}
\toprule[1.0pt]
                                       & MACs    & Encoding   Time  & Decoding   Time  \\ \hline
DCVC-DC w/ fp32            & 2642G   & 1005ms           & 765ms            \\ \hline
DCVC-FM w/ fp32                          & 2225G   & 1040ms           & 775ms            \\ \hline
DCVC-FM w/ fp16                          & 2225G   & 530ms            & 475ms            \\
\bottomrule[1.0pt]
\end{tabular}
}
\\
\vspace{0.2cm}
\raggedright\footnotesize{ \, Note:  Tested on NVIDIA 2080TI with using 1080p as input.}
\vspace{-0.3cm}
\label{tab_complexity}
\end{table}

\section{Conclusion and Limitation}
In conclusion, this paper proposes feature modulation techniques and resolves two major challenges limiting the practicality of NVC. Through a uniform quantization parameter sampling mechanism to help the learning of quantization scaler, we can finely modulate the latent feature and enable wide quality range support. Meanwhile, the rate control capability is demonstrated. We also have addressed issues with long prediction chains by modulating the temporal feature with a periodically refreshing mechanism. In addition, our DCVC-FM now supports both RGB and YUV colorspaces, and allows for low-precision inference. Our DCVC-FM represents a significant step in the evolution of NVC technology.

However, although DCVC-FM enables  low-precision float point inference, its speed is still far from real-time. In addition, the float point inference has cross-platform issue for the entropy coding in NVC.  In the future, we will investigate these topics and build a more powerful NVC which can be widely deployed in practical products.

{
    \small
    %\bibliographystyle{ieeenat_fullname}
    %\bibliography{main}

\input{main_arxiv.bbl}
}

\clearpage
\begin{appendices}
This document provides the supplementary material to our proposed neural video codec (NVC), i.e. DCVC-FM.

\section{Test Settings}
For a thorough comparative analysis, we compare the NVCs and traditional codecs in both YUV420 and RGB colorspaces. 

\textbf{YUV420 colorspace.} Most traditional codecs and practical applications mainly adopt YUV420 colorspace as input and output, and are optimized in this colorspace. Thus, the comparison between NVC and traditional codec in YUV420  colorspace is quite important for evaluating the development progress of NVC. 
For traditional codec,  HM \cite{HM}, VTM \cite{VTM}, and ECM \cite{ECM} are tested, where HM, VTM, ECM are the reference software of H.265, H.266, and the under-developing next generation traditional codec, respectively. For the three traditional codecs,    \textit{encoder\_lowdelay\_main10.cfg}, \textit{encoder\_lowdelay\_vtm.cfg}, and  \textit{encoder\_lowdelay\_ecm.cfg} config files are used, respectively. The parameters for each video are as:
\begin{itemize}
	\item % {\bf Settings for YUV444}\par 
	-c \{{\em config file name}\}\par
	-\/-InputFile=\{{\em input video name}\}\par
	-\/-InputBitDepth=8\par
	-\/-OutputBitDepth=8 \par
	-\/-OutputBitDepthC=8 \par
	-\/-FrameRate=\{{\em frame rate}\}\par
	-\/-DecodingRefreshType=2\par
	-\/-FramesToBeEncoded=\{{\em frame number}\}\par
	-\/-SourceWidth=\{{\em width}\}\par
	-\/-SourceHeight=\{{\em height}\}\par
	-\/-IntraPeriod=\{{\em intra period}\}\par
	-\/-QP=\{{\em qp}\}\par
	-\/-Level=6.2\par
	-\/-BitstreamFile=\{{\em bitstream file name}\}\par
\end{itemize}

\textbf{RGB colorspace.}  As the raw formats of all testsets are in YUV420 colorspace. Thus, to test RGB video, we need to convert them from YUV420 to RGB colorspace. We follow JPEG AI \cite{jpeg_ai,jpeg_ai2} and \cite{li2023neural}, and  use BT.709 to convert the raw YUV420 video to RGB video. This is because
using BT.709 obtains higher compression ratio under the similar visual quality when compared with the commonly-used  BT.601. \cite{li2023neural} shows, when traditional codecs test RGB videos, using 10-bit YUV444 as the internal colorspace  achieves better compression ratio than directly using RGB, although the final distortion is measured in RGB. So we also follow this setting.
For HM, VTM, and ECM,  \textit{encoder\_lowdelay\_main\_rext.cfg}, \textit{encoder\_lowdelay\_vtm.cfg}, and  \textit{encoder\_lowdelay\_ecm.cfg} config files are used, respectively.  The parameters for each video are as:
\begin{itemize}
	\item % {\bf Settings for YUV444}\par 
	-c \{{\em config file name}\}\par
	-\/-InputFile=\{{\em input file name}\}\par
	-\/-InputBitDepth=10\par
	-\/-OutputBitDepth=10 \par
	-\/-OutputBitDepthC=10 \par
	-\/-InputChromaFormat=444\par
	-\/-FrameRate=\{{\em frame rate}\}\par
	-\/-DecodingRefreshType=2\par
	-\/-FramesToBeEncoded=\{{\em frame number}\}\par
	-\/-SourceWidth=\{{\em width}\}\par
	-\/-SourceHeight=\{{\em height}\}\par
	-\/-IntraPeriod=\{{\em intra period}\}\par
	-\/-QP=\{{\em qp}\}\par
	-\/-Level=6.2\par
	-\/-BitstreamFile=\{{\em bitstream file name}\}\par
\end{itemize}

It is noted that, for both YUV420 and RGB colorspaces, all coding tools and reference structure of traditional codecs use their best settings to represent their  best compression ratio.

\section{Temporal Feature Modulation}
In this paper, we specially design the feature refresh mechanism to modulate the temporal feature. It will improve the effectiveness of feature propagation. The default refresh period is set as 32. Here we test different refresh period settings, and the corresponding comparisons are in Table \ref{aba_refresh}, where the refresh period 0 means disabling the feature refresh. From this table, we can see that the bitrate saving initially increases with the refresh period size. When the  refresh period size is larger than 32 (i.e. 64 and 96), the performance begins to decay. This phenomenon shows that both too small or too large refresh period sizes are not proper. If period size is  too small, the addition bitrate cost is too heavy because the frame with  feature refresh will not only have better quality but also have a larger bitrate cost. 
Conversely, an excessively large period size can lead to the propagation of features contaminated by the accumulated errors or containing higher amounts of uncorrelated information.
So currently the refresh period 32 is a good trade-off. Nevertheless, it should be noted that the best refresh period size may vary across different videos. A content-adaptive approach to determine the refresh period size would likely yield better results. We will investigate it in the future.

\begin{table}[t]
	\centering
	\caption{BD-Rate (\%) of using different feature refresh periods.}
	\scalebox{1.1}{
		\renewcommand{\arraystretch}{1.0}
		\setlength{\tabcolsep}{4pt}
		\small
		\begin{tabular}{ccccccc}
			\toprule[1.0pt]
			Refresh period &   0    &8       & 16         & 32          & 64       & 96        \\ \hline
			BD-Rate &   0    &--7.2    & --17.1      & --20.0       & --17.9     & --14.8        \\
			
			\bottomrule[1.0pt]
		\end{tabular}
	}
	\vspace{-0.0cm}
	\label{aba_refresh}
\end{table}

\section{Rate Control}
We implement a simple rate control algorithm to just showcase the feasibility by adjusting the quantization parameter $q_t$ values in the model. Because of the bitrate fluctuation among frames, we only adjust the quantization parameter $q_t$ at even frames. The rate control algorithm is provided in Algorithm \ref{alg:rc}. It should be noted that this algorithm is a simple implementation  to demonstrate the feasibility of rate control. More advanced rate control algorithms can be designed based on our codec and we will investigate it in the future. 

\begin{algorithm}
	\caption{Rate Control Algorithm}\label{alg:rc}
	\textbf{Input:} \\
	\hspace*{\algorithmicindent} $cbs$: current buffer size (set as 0 for the first frame) \\
	\hspace*{\algorithmicindent} $tbs$: target buffer size (set as 0 for the first frame) \\
	\hspace*{\algorithmicindent} $q$: current q value (set as 32 for the first frame) \\
	\hspace*{\algorithmicindent} $cfs$: current frame size \\
	\hspace*{\algorithmicindent} $afs$: average frame size \\
	\hspace*{\algorithmicindent} $fidx$: current frame index \\
	\textbf{Output} \\
	\hspace*{\algorithmicindent} $cbs$: updated buffer size for the next frame \\
	\hspace*{\algorithmicindent} $tbs$: updated target buffer size for the next frame  \\
	\hspace*{\algorithmicindent} $q$: updated q value for the next frame \\
	\begin{algorithmic}
		\State $cbs += cfs$
		\State $cbs -= afs$
		\If{$fidx \mod 2 == 1$}
		\State \Return $cbs$, $tbs$, $q$
		\EndIf
		
		\State $buff\_diff = cbs - tbs$
		\State $tbs = cbs * 0.95$
		\If{$buff\_diff > 0$}
		\If{$cbs > 10*cfs$}
		\State $q -= 12$
		\ElsIf{$cbs > 5*cfs$}
		\State $q -= 6$
		\ElsIf{$cbs > 2 * cfs$}
		\State $q -= 2$
		\ElsIf{$buff\_diff > 0.5 * cfs \And cbs > -cfs$}
		\State $q -= 1$
		\EndIf
		\ElsIf{$buff\_diff < 0$}
		\If{$cbs < -10*cfs$}
		\State $q += 12$
		\ElsIf{$cbs < -5*cfs$}
		\State $q += 6$
		\ElsIf{$cbs < -2 * cfs$}
		\State $q += 2$
		\ElsIf{$buff\_diff < -0.5 * cfs \And cbs < cfs$}
		\State $q += 1$
		\EndIf
		\EndIf
		\State $q=clip(0, 63, q)$
		\State \Return $cbs$, $tbs$, $q$
	\end{algorithmic}
\end{algorithm}

\section{Reimplemented \textit{grid\_sample}}
%The attached codes for the reimplemented \textit{grid\_sample} function are in the attached ``grid\_sample\_reimplement" folder. Please read the ``README.md" file first and then test the code in ``test\_mc.py".  As shown in ``test\_mc.py", we also calculate the error ratio. 
The reimplemented \textit{grid\_sample} function can be found in our release codes. We also compare the error ratio for the randomly generated feature and motion vector. The error ratio is as high as 16.149\%  if we directly feed the 16-bit tensor value to the default \textit{grid\_sample} function. By contrast, the error ratio is 0.026\% if using our reimplemented and improved \textit{grid\_sample} function.

\section{Rate-Distortion Curves}
\begin{figure*}[t]
	
	\minipage{0.33\textwidth}
	\includegraphics[width=1.07\linewidth,height=0.9\linewidth]{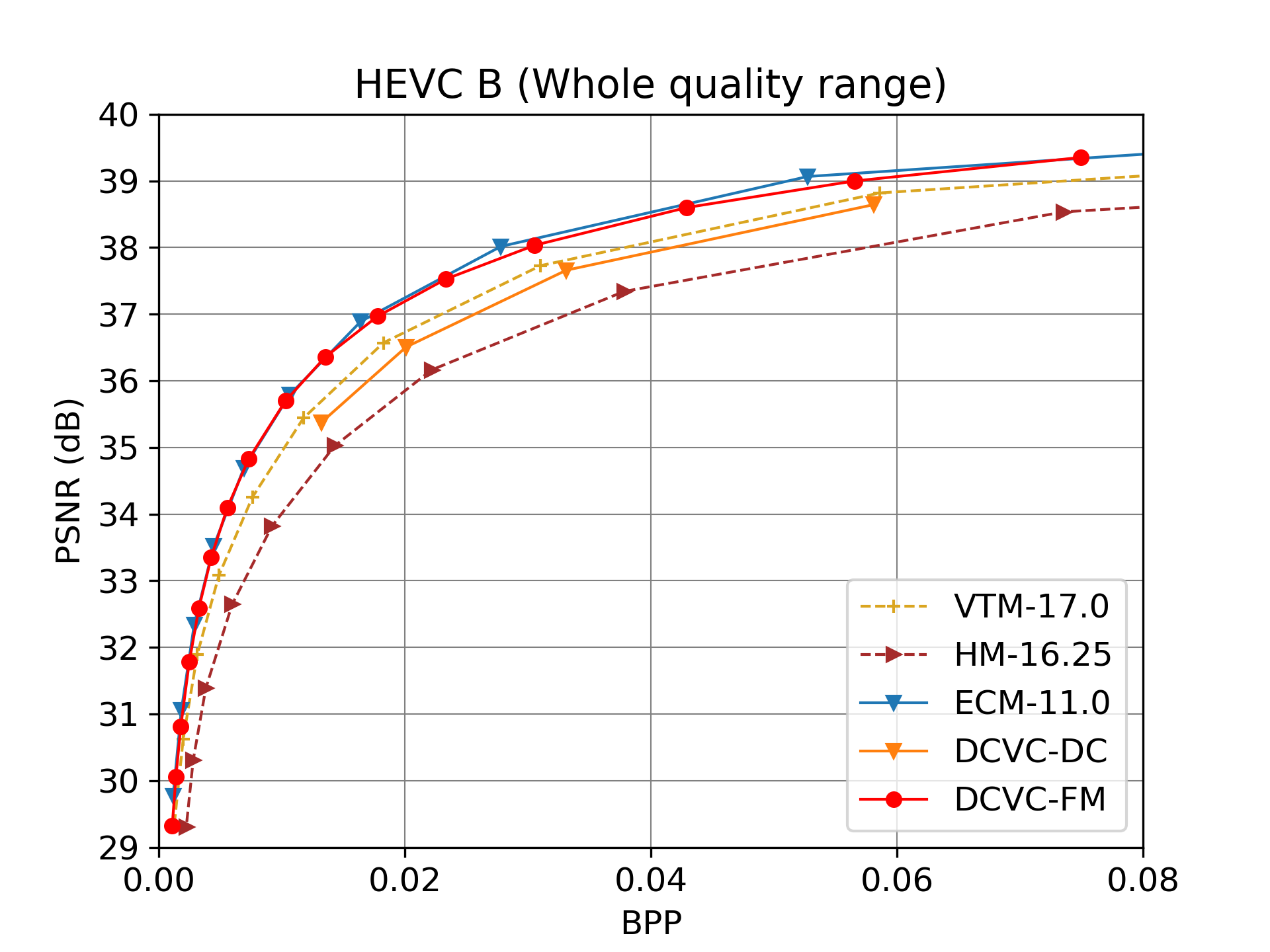}
	\endminipage
	\minipage{0.33\textwidth}
	\includegraphics[width=1.07\linewidth,height=0.9\linewidth]{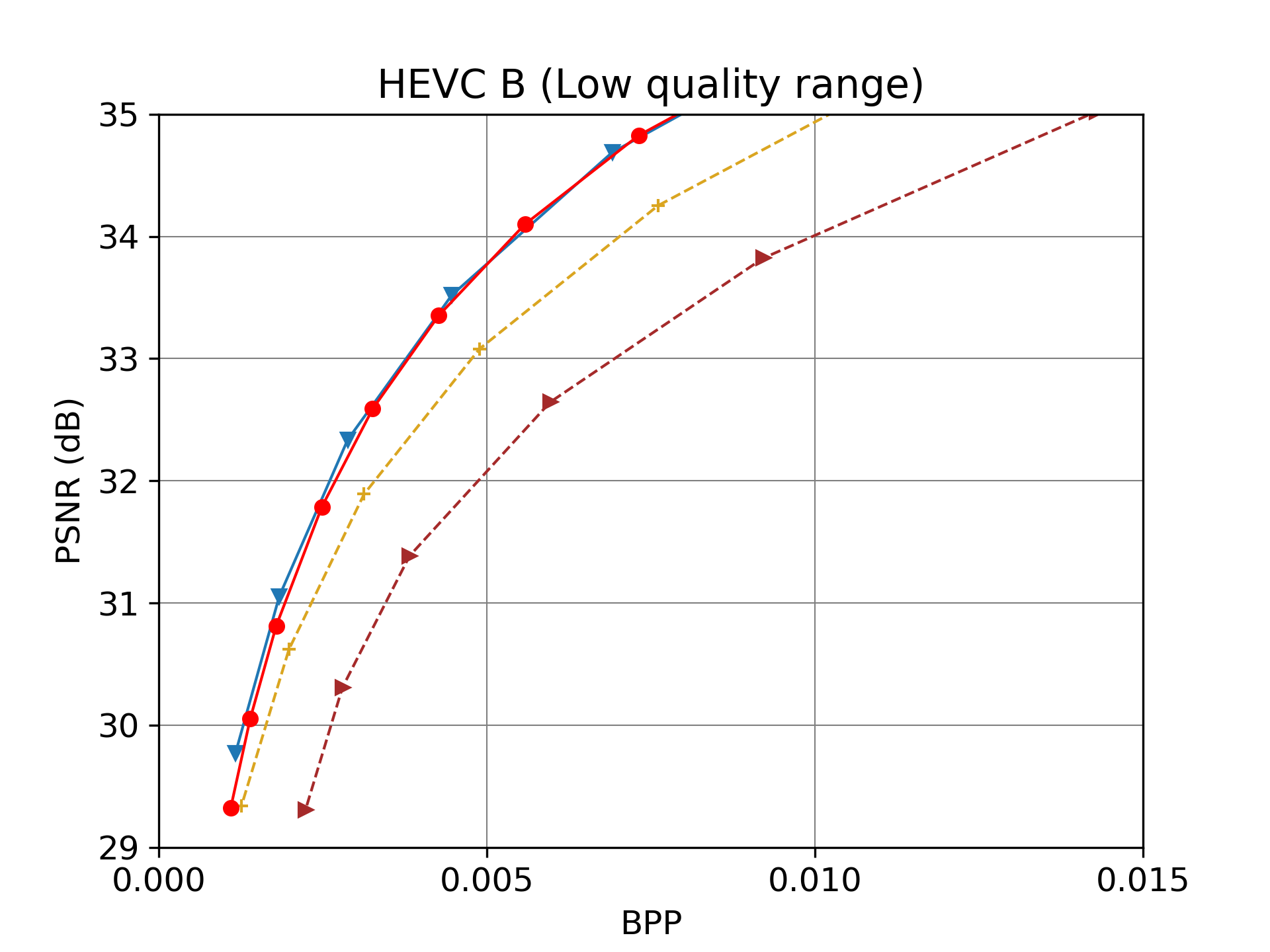}
	\endminipage
	\minipage{0.33\textwidth}%
	\includegraphics[width=1.07\linewidth,height=0.9\linewidth]{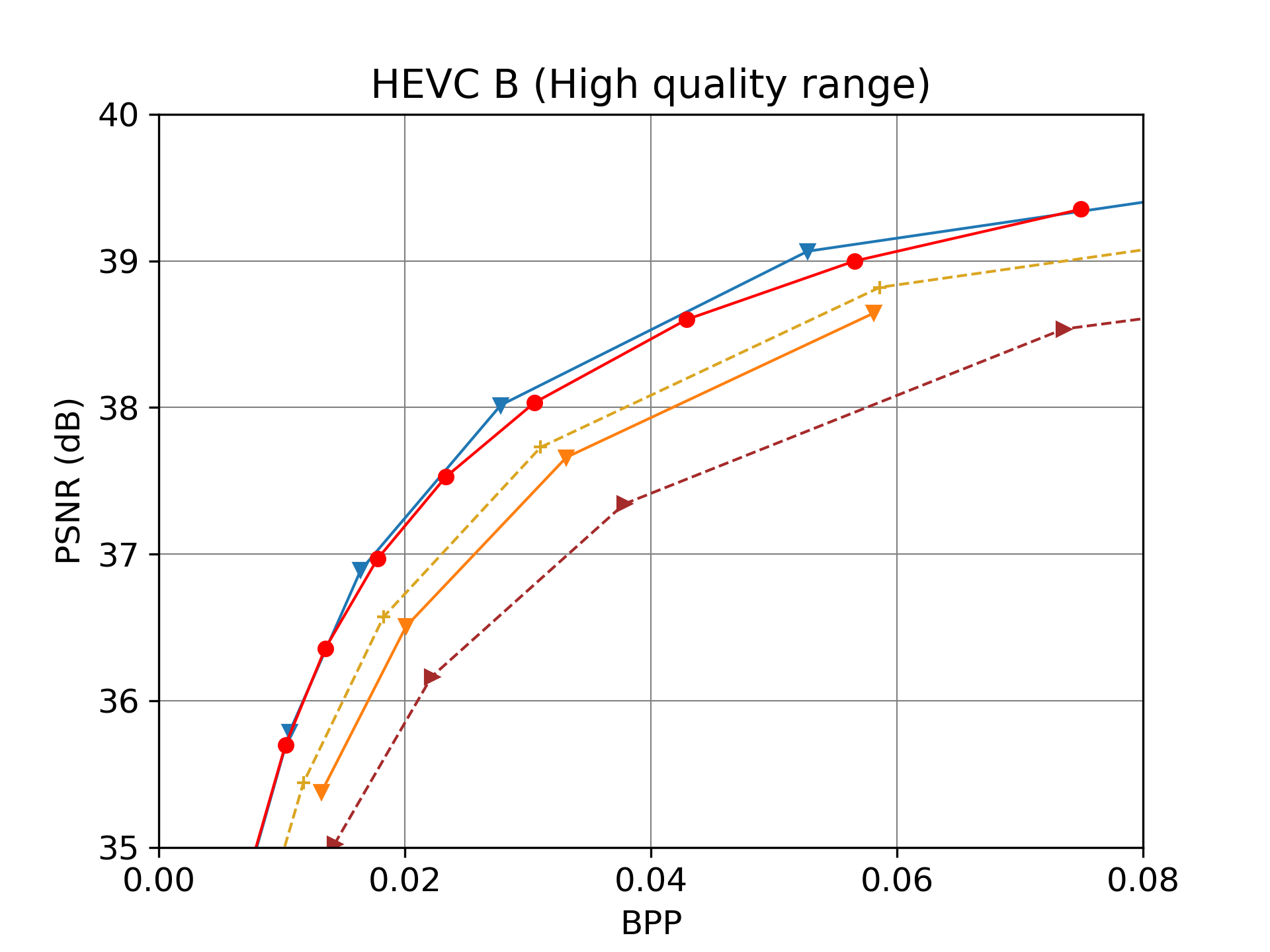}
	\endminipage
	
	\minipage{0.33\textwidth}
	\includegraphics[width=1.07\linewidth,height=0.9\linewidth]{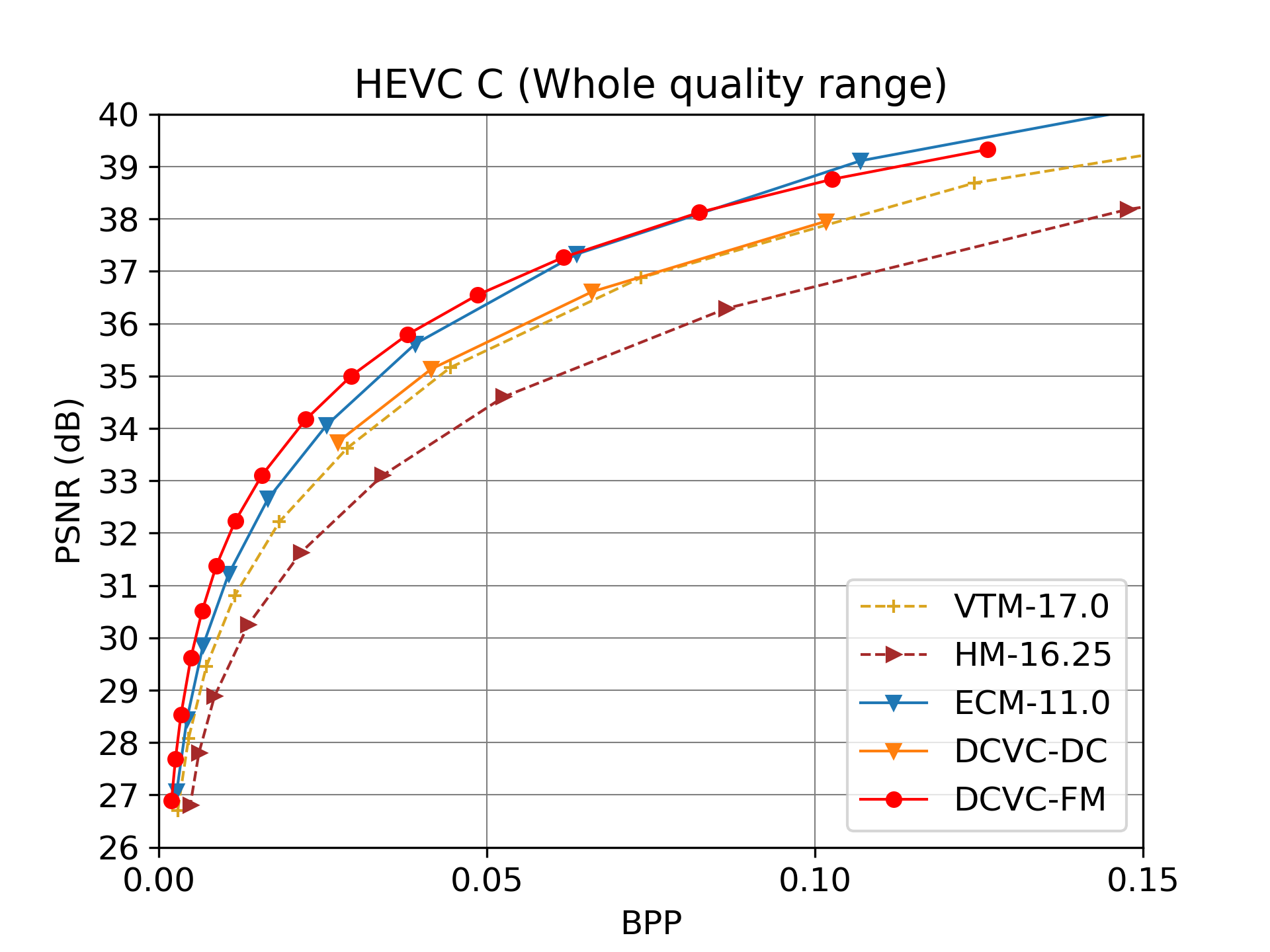}
	\endminipage
	\minipage{0.33\textwidth}
	\includegraphics[width=1.07\linewidth,height=0.9\linewidth]{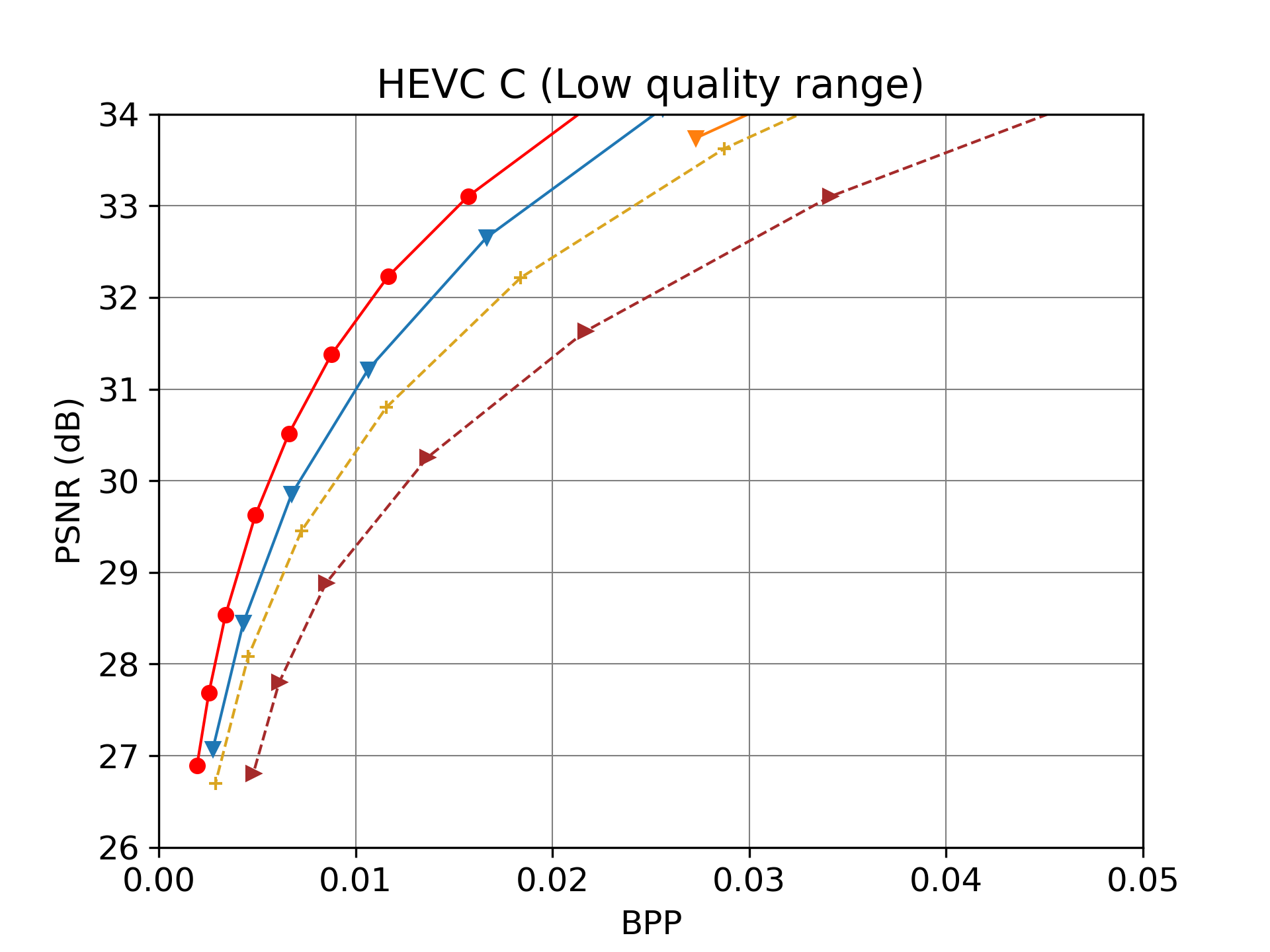}
	\endminipage
	\minipage{0.33\textwidth}%
	\includegraphics[width=1.07\linewidth,height=0.9\linewidth]{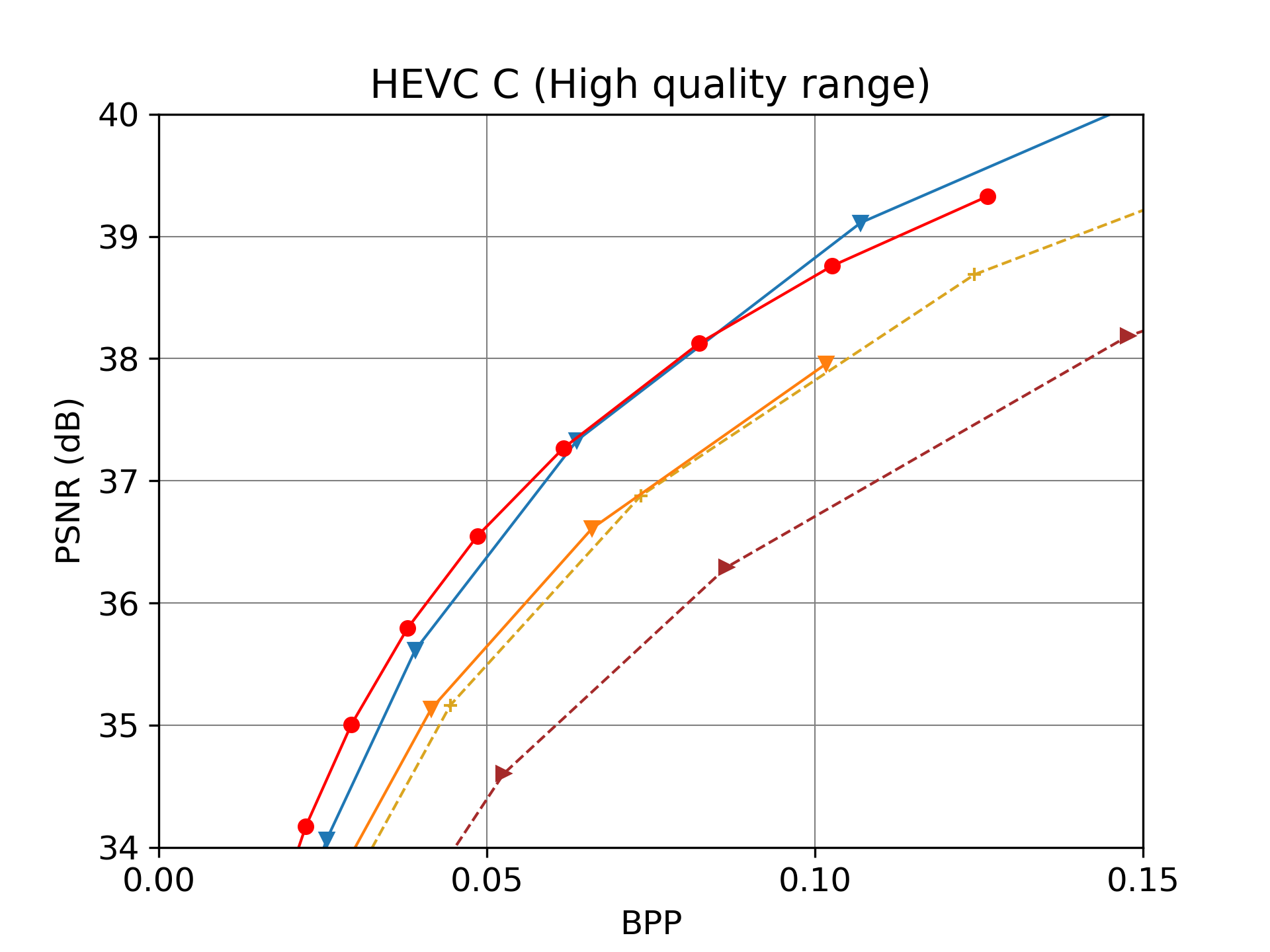}
	\endminipage
	
	\minipage{0.33\textwidth}
	\includegraphics[width=1.07\linewidth,height=0.9\linewidth]{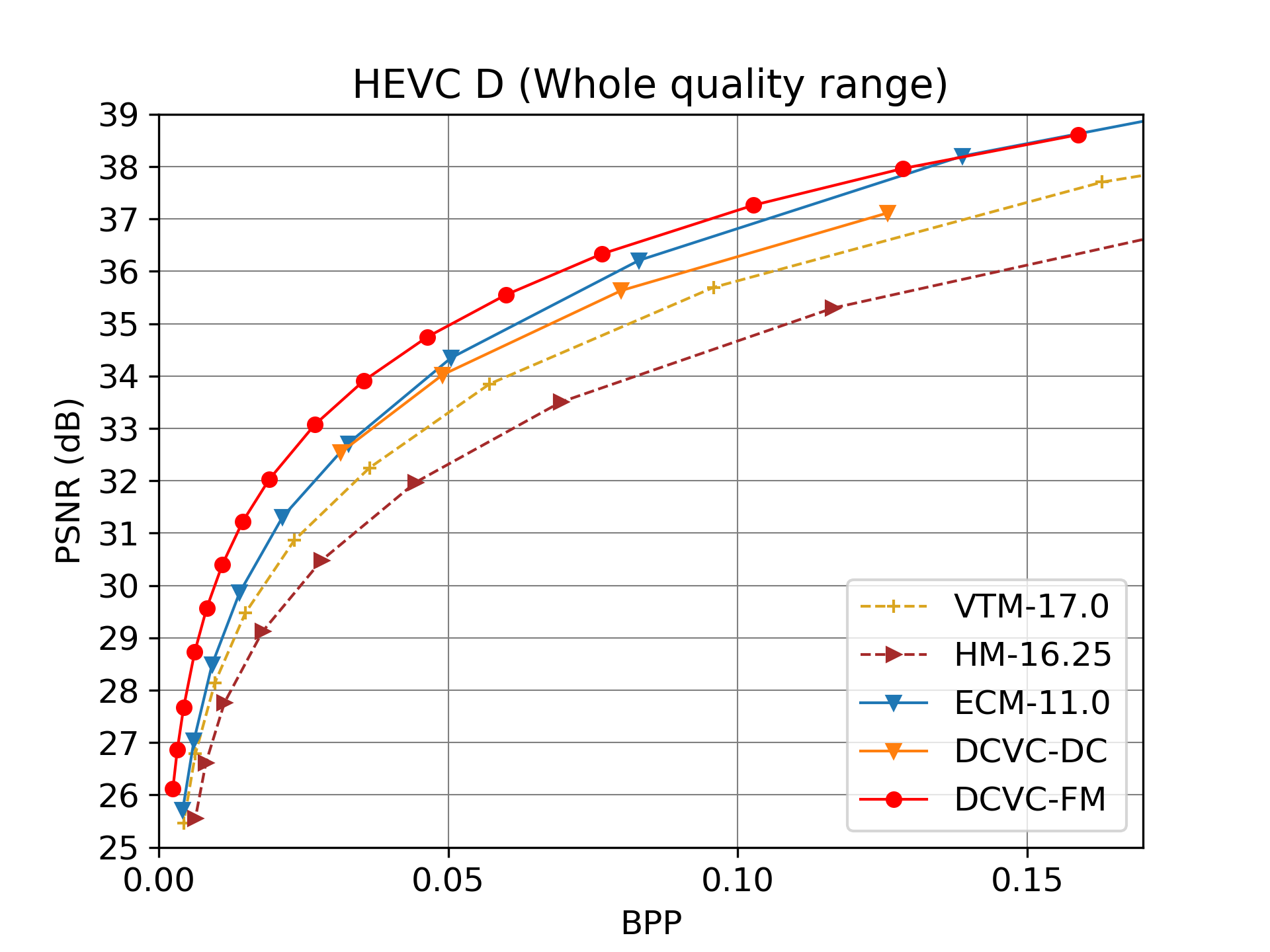}
	\endminipage
	\minipage{0.33\textwidth}
	\includegraphics[width=1.07\linewidth,height=0.9\linewidth]{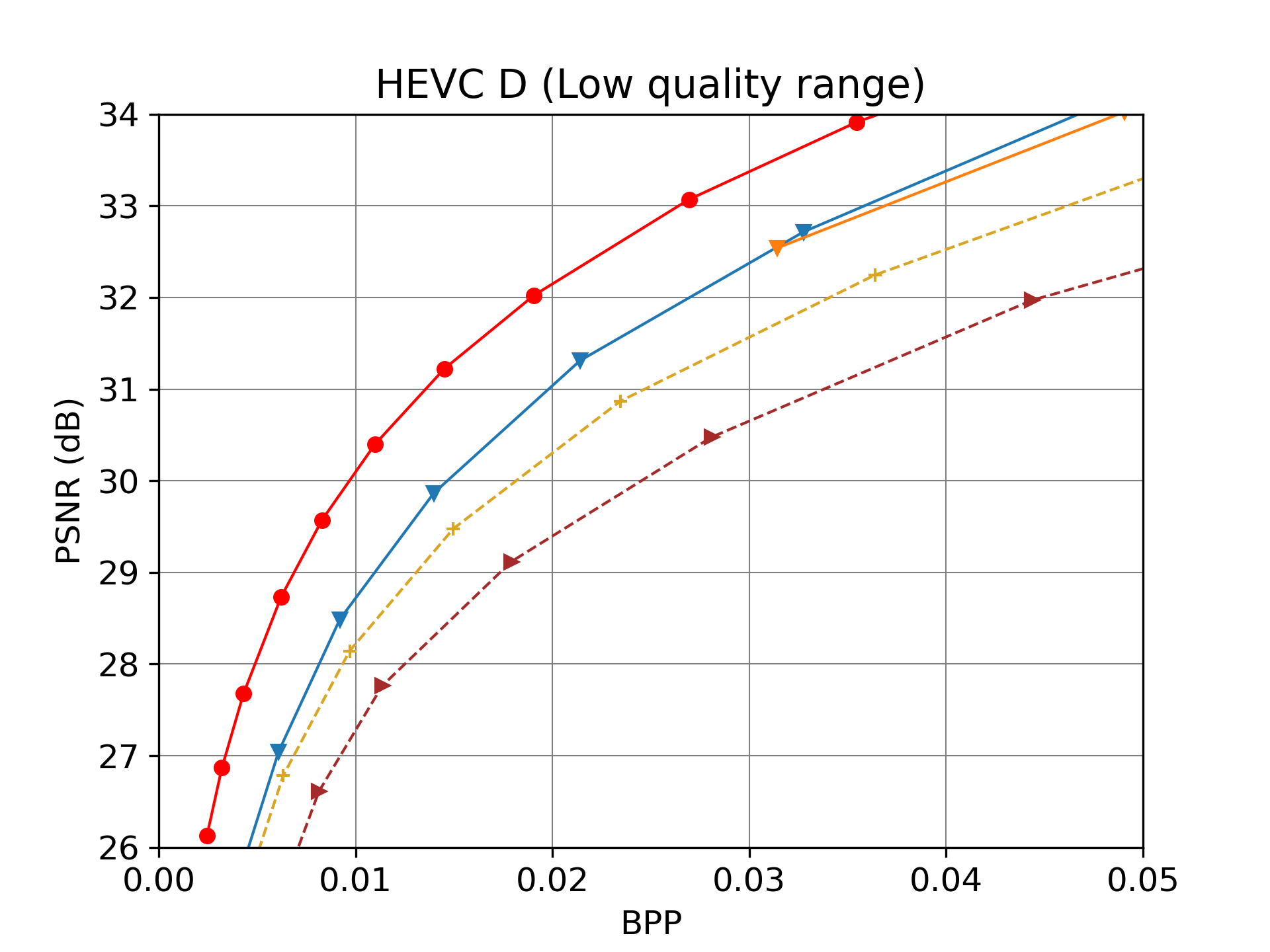}
	\endminipage
	\minipage{0.33\textwidth}%
	\includegraphics[width=1.07\linewidth,height=0.9\linewidth]{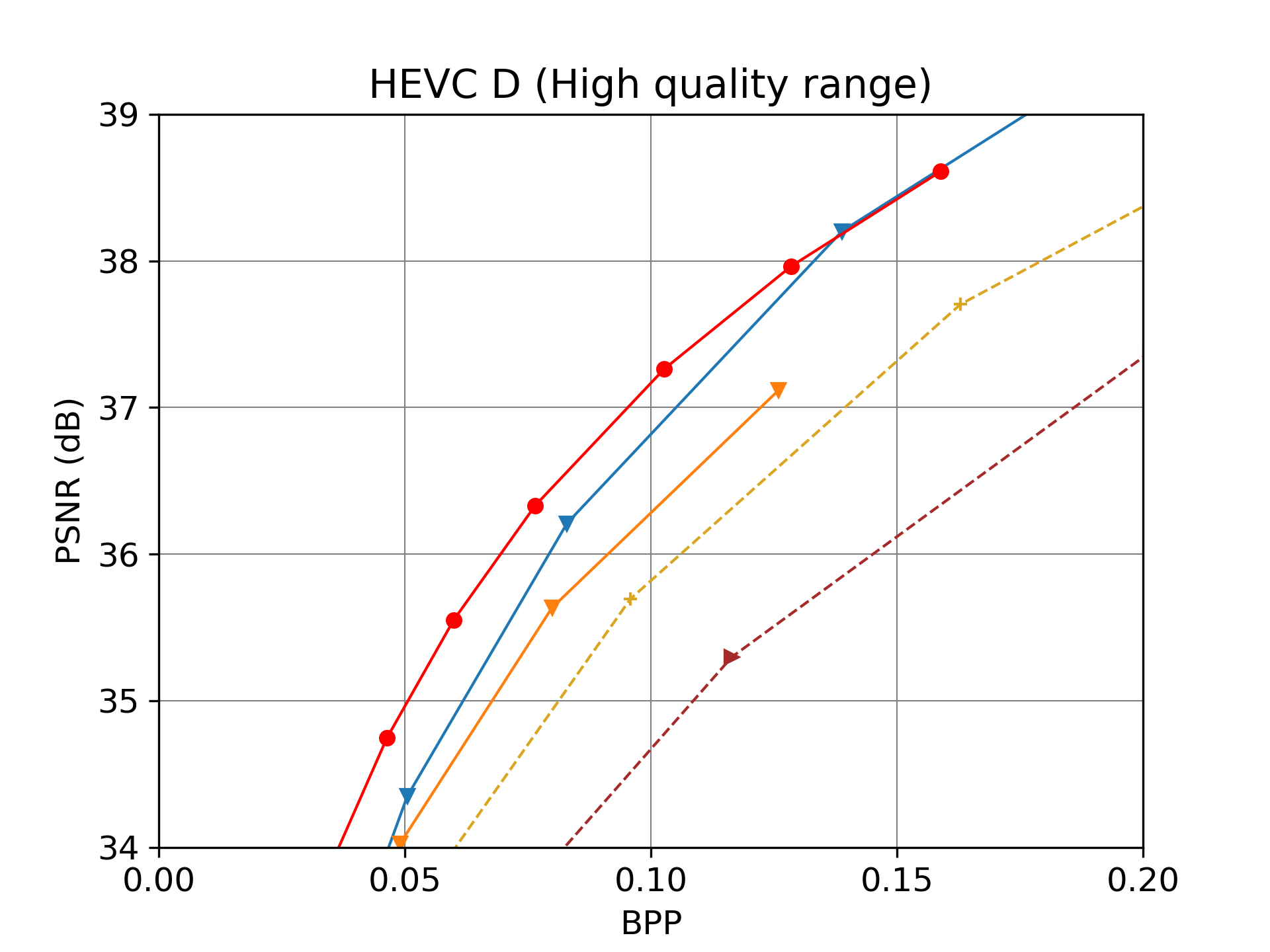}
	\endminipage
	\vspace{-4mm}
	\caption{Rate and distortion curves for HEVC B, HEVC C, and HEVC D datasets. Each row shows a dataset.
		From left to right the figures are overall quality range, relatively low quality range and relatively high quality range, respectively. The comparison is in YUV420 colorspace. All frames with intra-period = --1.  }
	\vspace{-4mm}
	\label{fig_yuv_psnr_allf_curve_part1}
\end{figure*}

\begin{figure*}[t]
	
	\minipage{0.33\textwidth}
	\includegraphics[width=1.07\linewidth,height=0.9\linewidth]{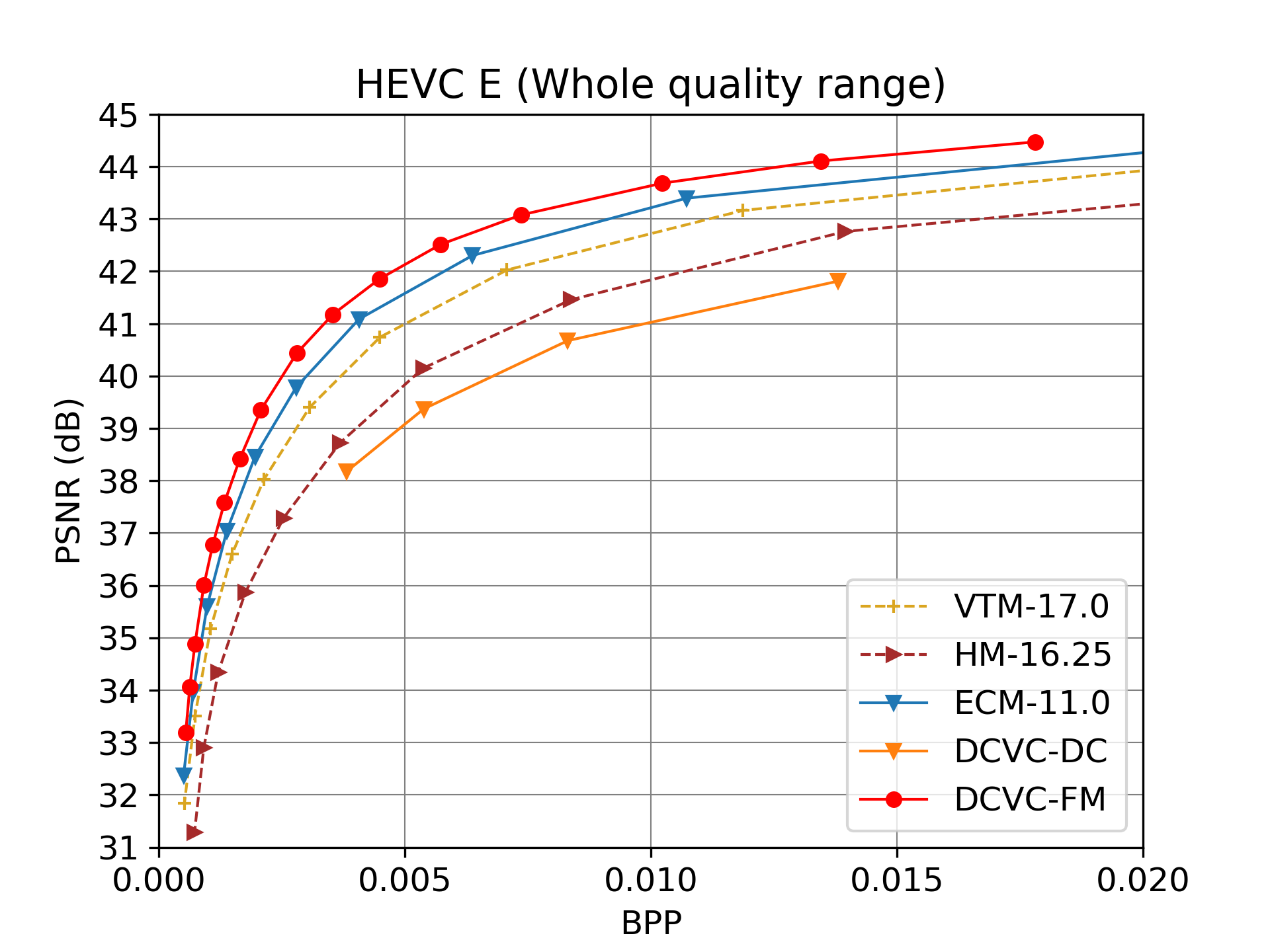}
	\endminipage
	\minipage{0.33\textwidth}
	\includegraphics[width=1.07\linewidth,height=0.9\linewidth]{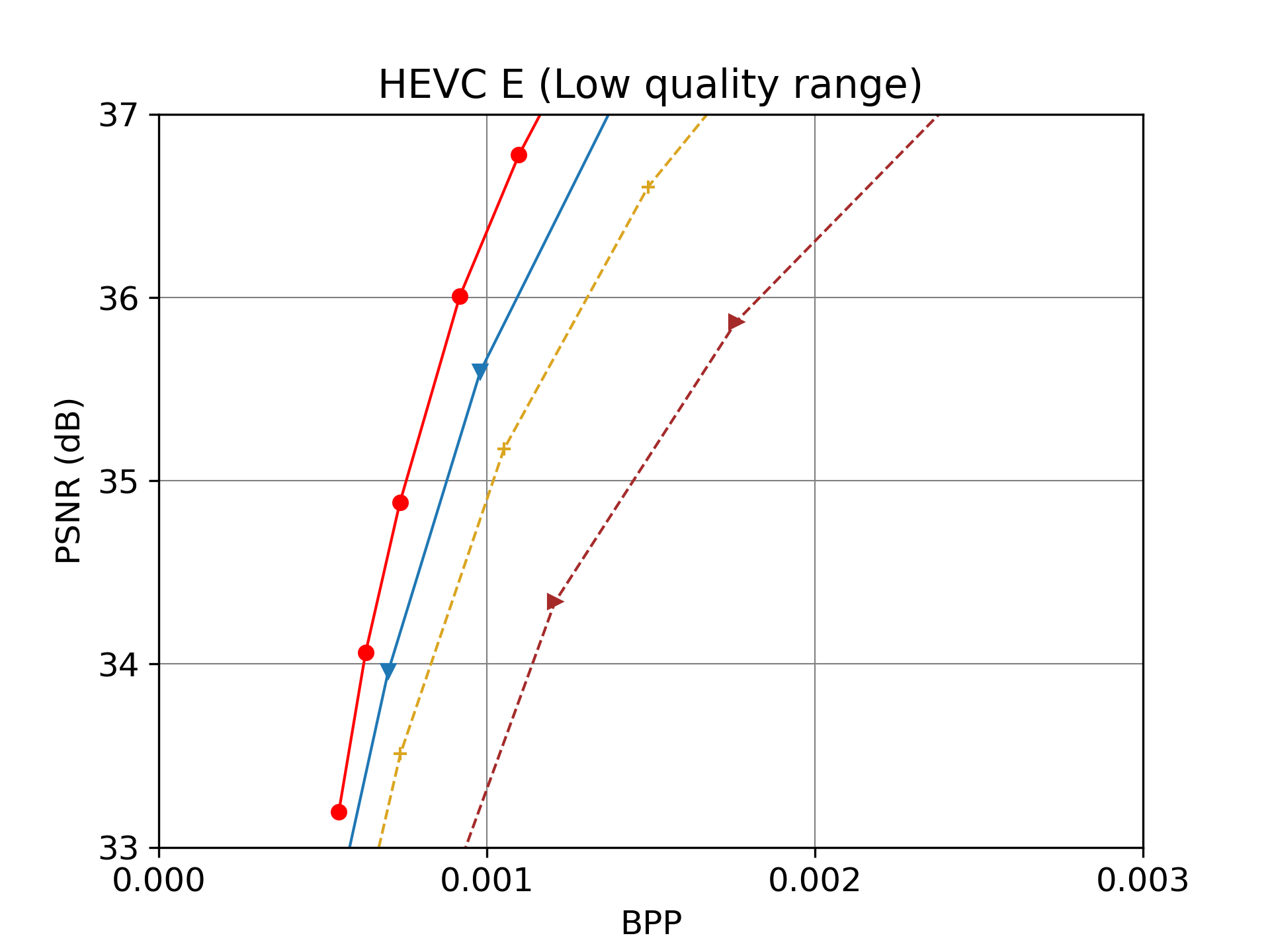}
	\endminipage
	\minipage{0.33\textwidth}%
	\includegraphics[width=1.07\linewidth,height=0.9\linewidth]{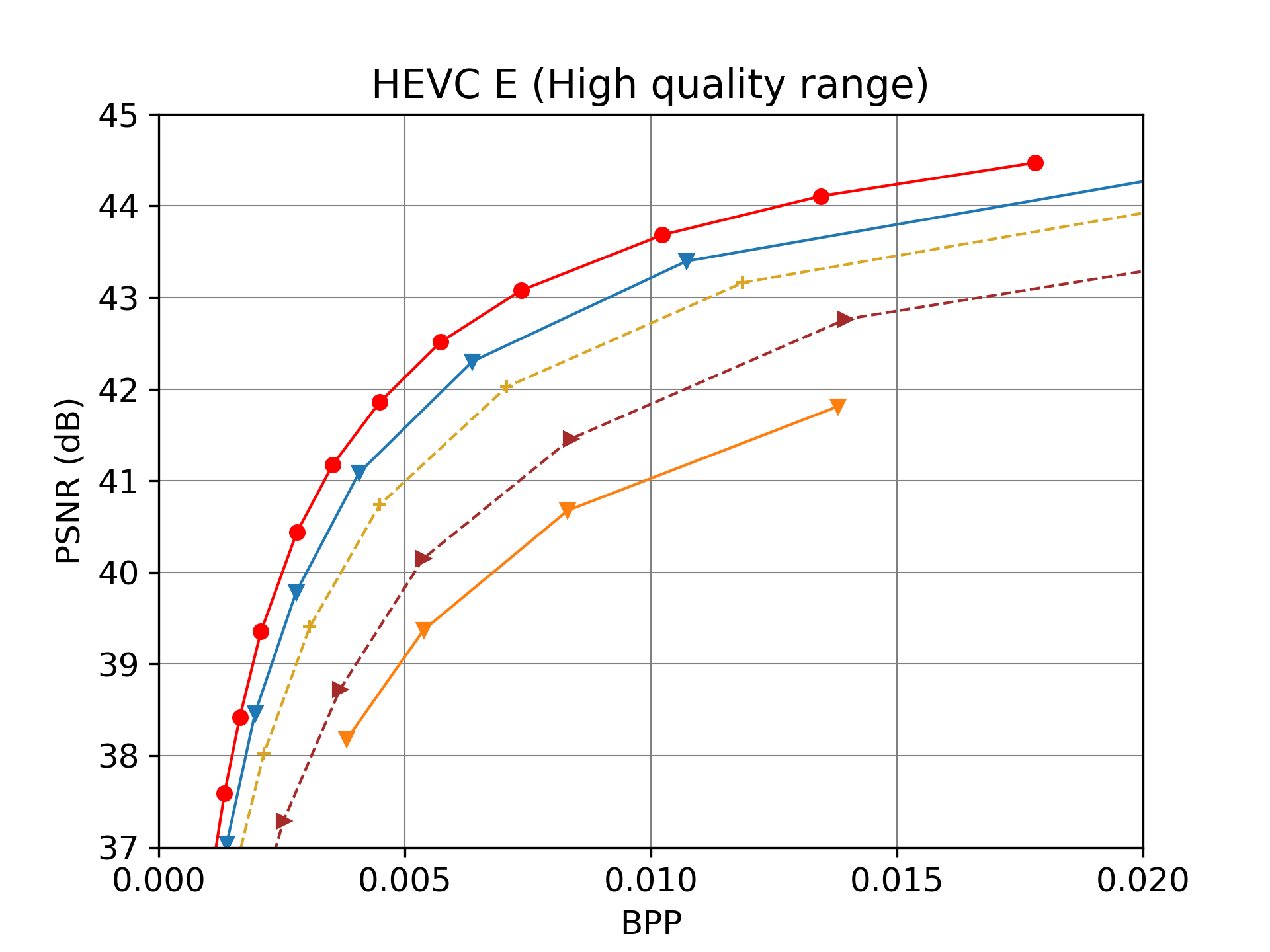}
	\endminipage
	
	\minipage{0.33\textwidth}
	\includegraphics[width=1.07\linewidth,height=0.9\linewidth]{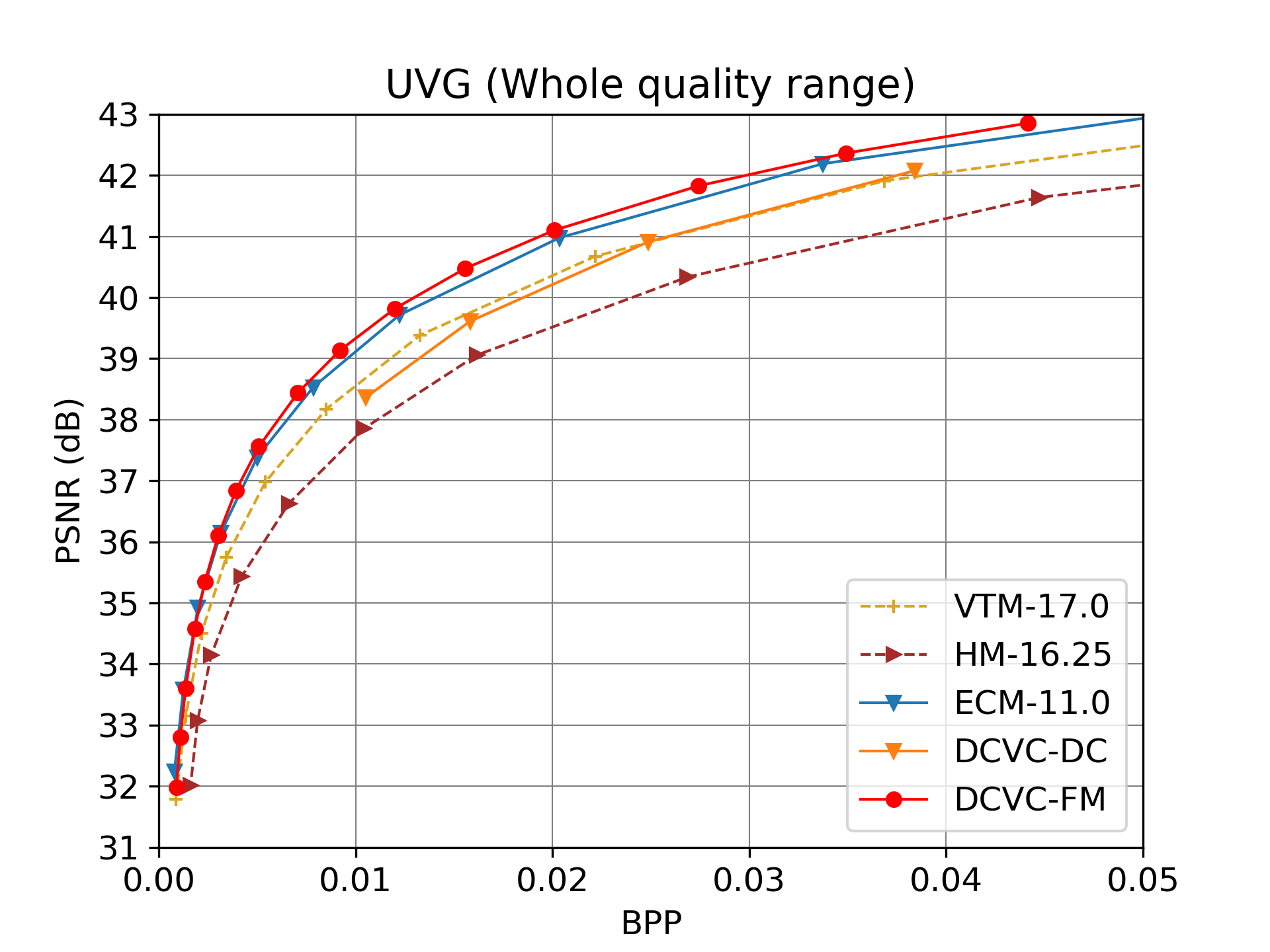}
	\endminipage
	\minipage{0.33\textwidth}
	\includegraphics[width=1.07\linewidth,height=0.9\linewidth]{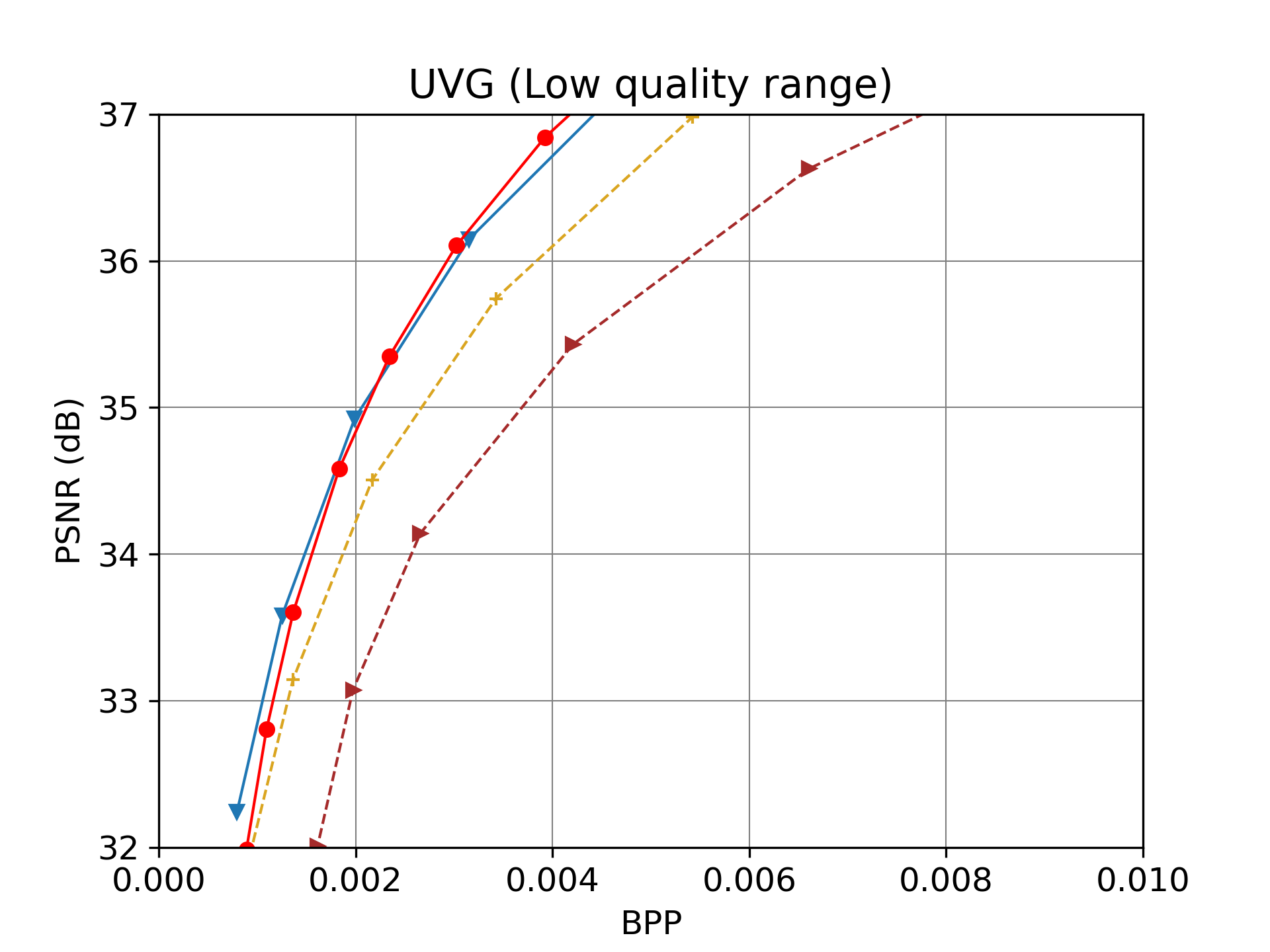}
	\endminipage
	\minipage{0.33\textwidth}%
	\includegraphics[width=1.07\linewidth,height=0.9\linewidth]{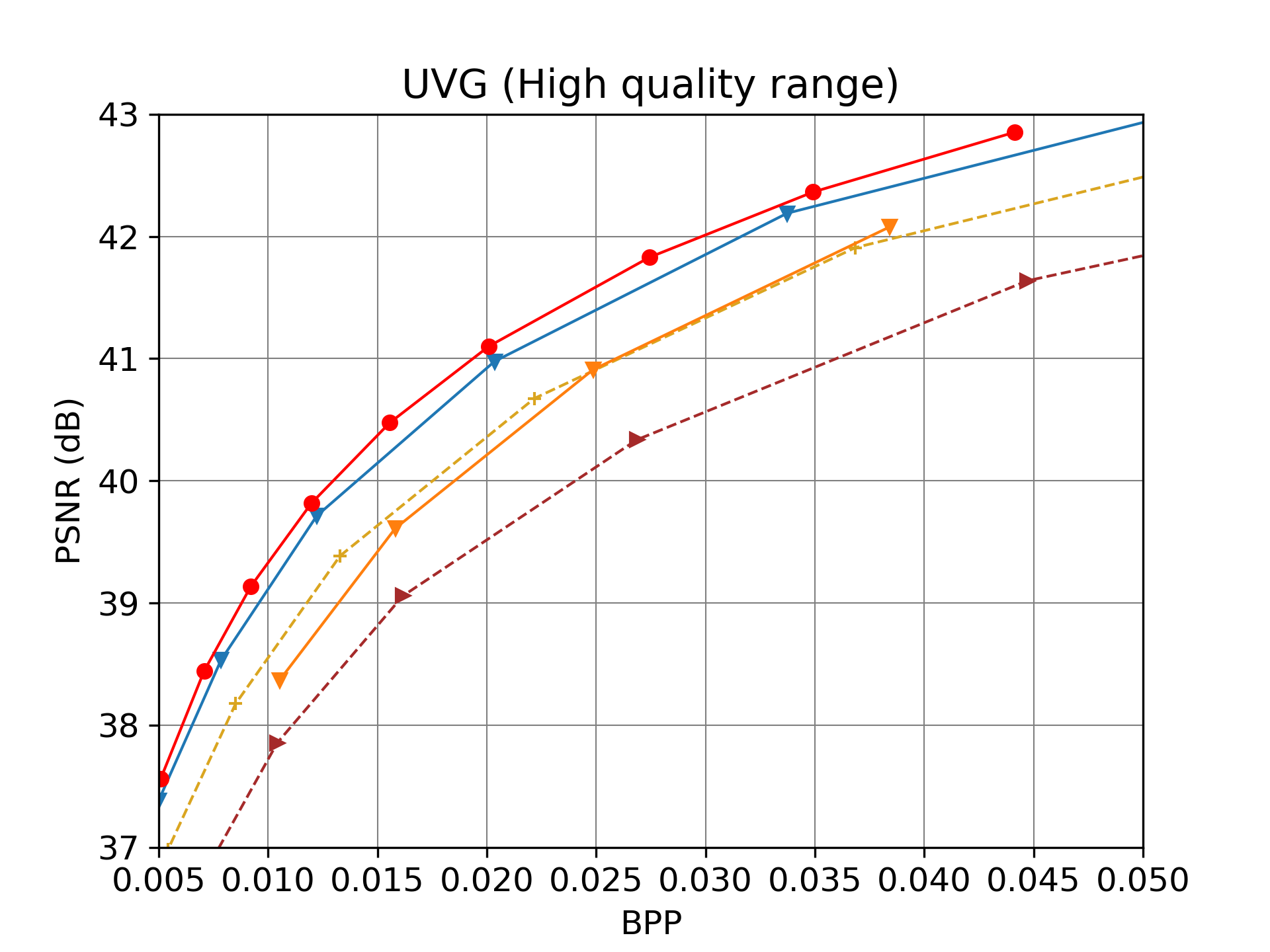}
	\endminipage
	
	\minipage{0.33\textwidth}
	\includegraphics[width=1.07\linewidth,height=0.9\linewidth]{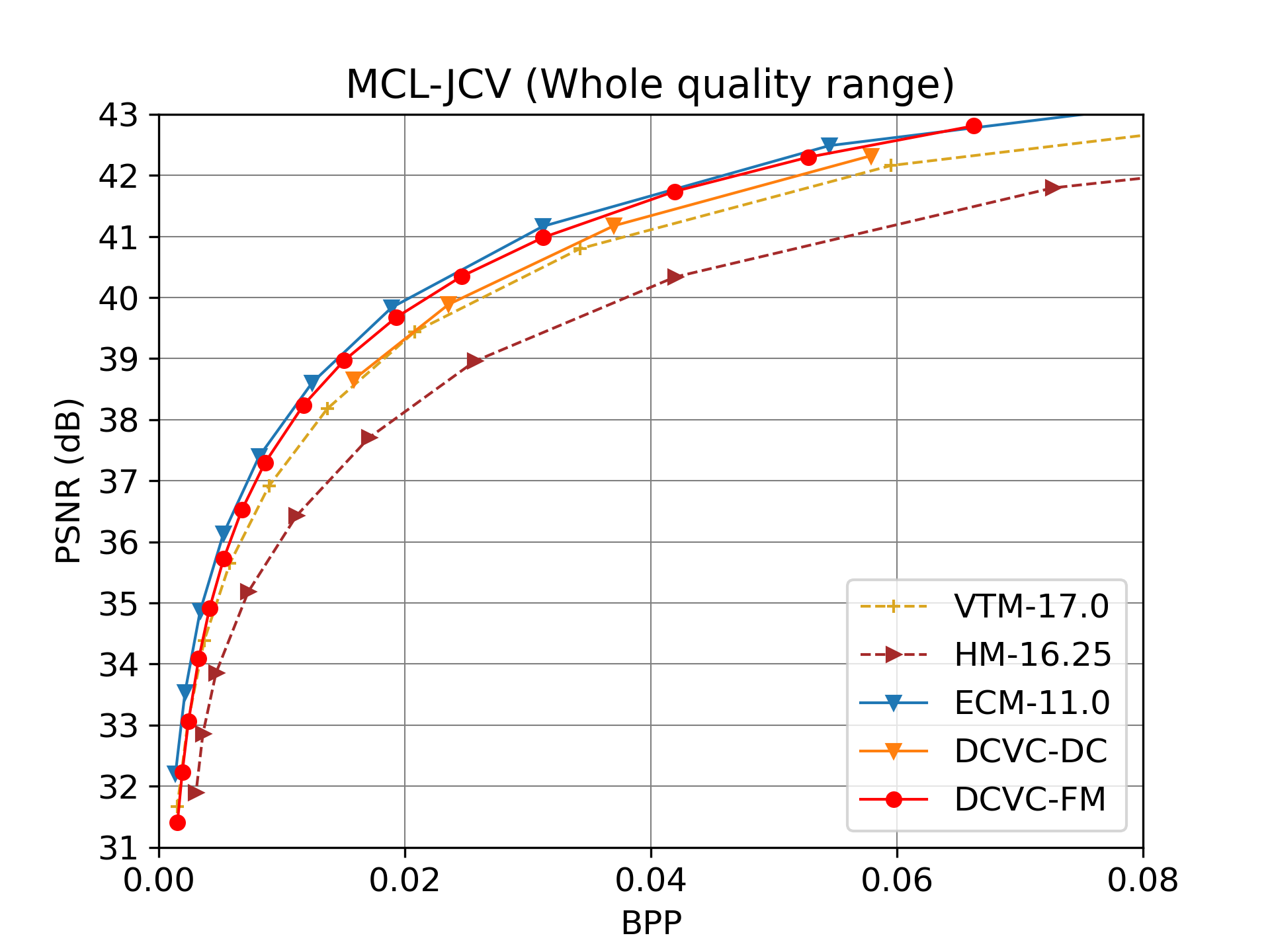}
	\endminipage
	\minipage{0.33\textwidth}
	\includegraphics[width=1.07\linewidth,height=0.9\linewidth]{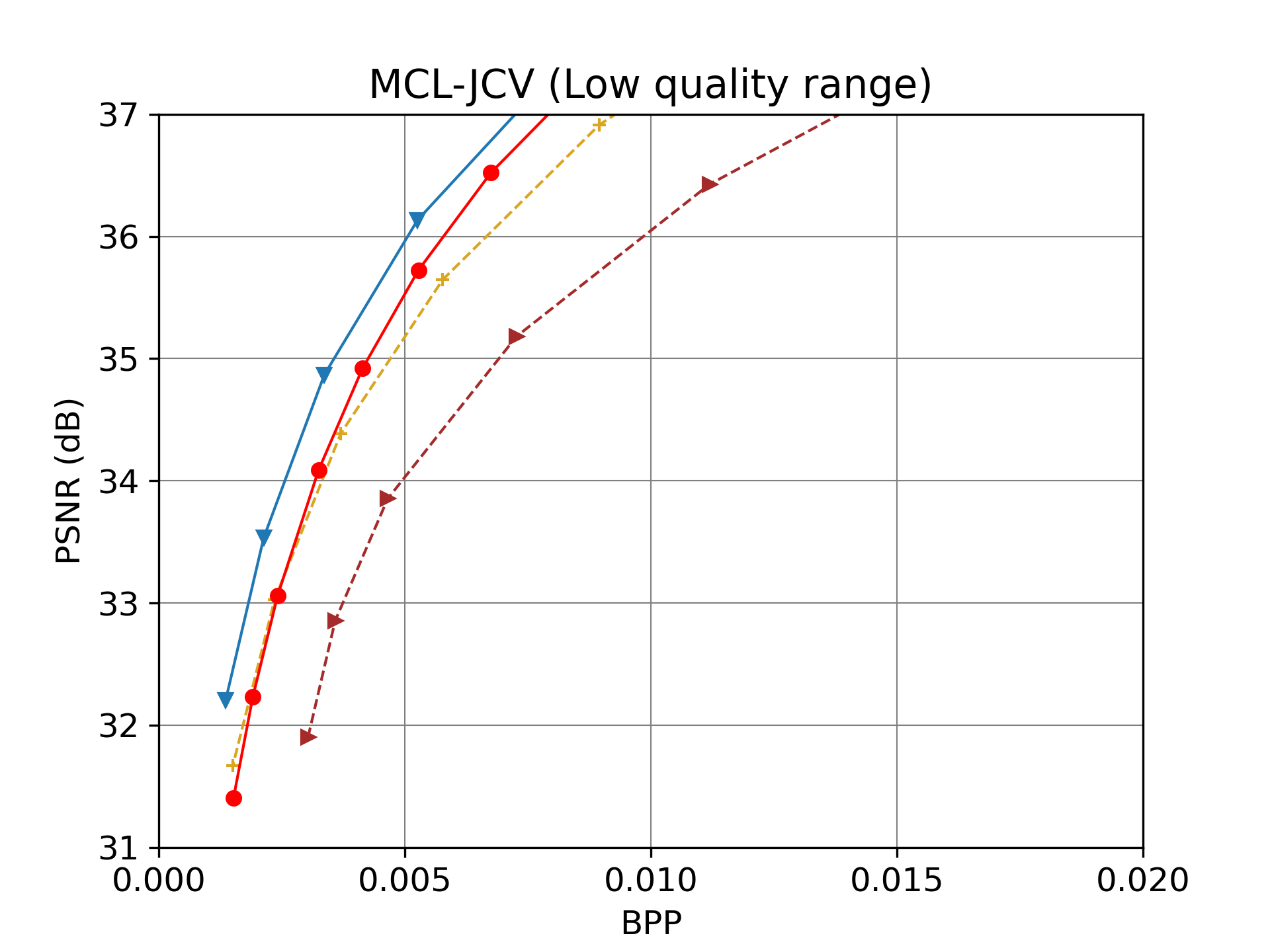}
	\endminipage
	\minipage{0.33\textwidth}%
	\includegraphics[width=1.07\linewidth,height=0.9\linewidth]{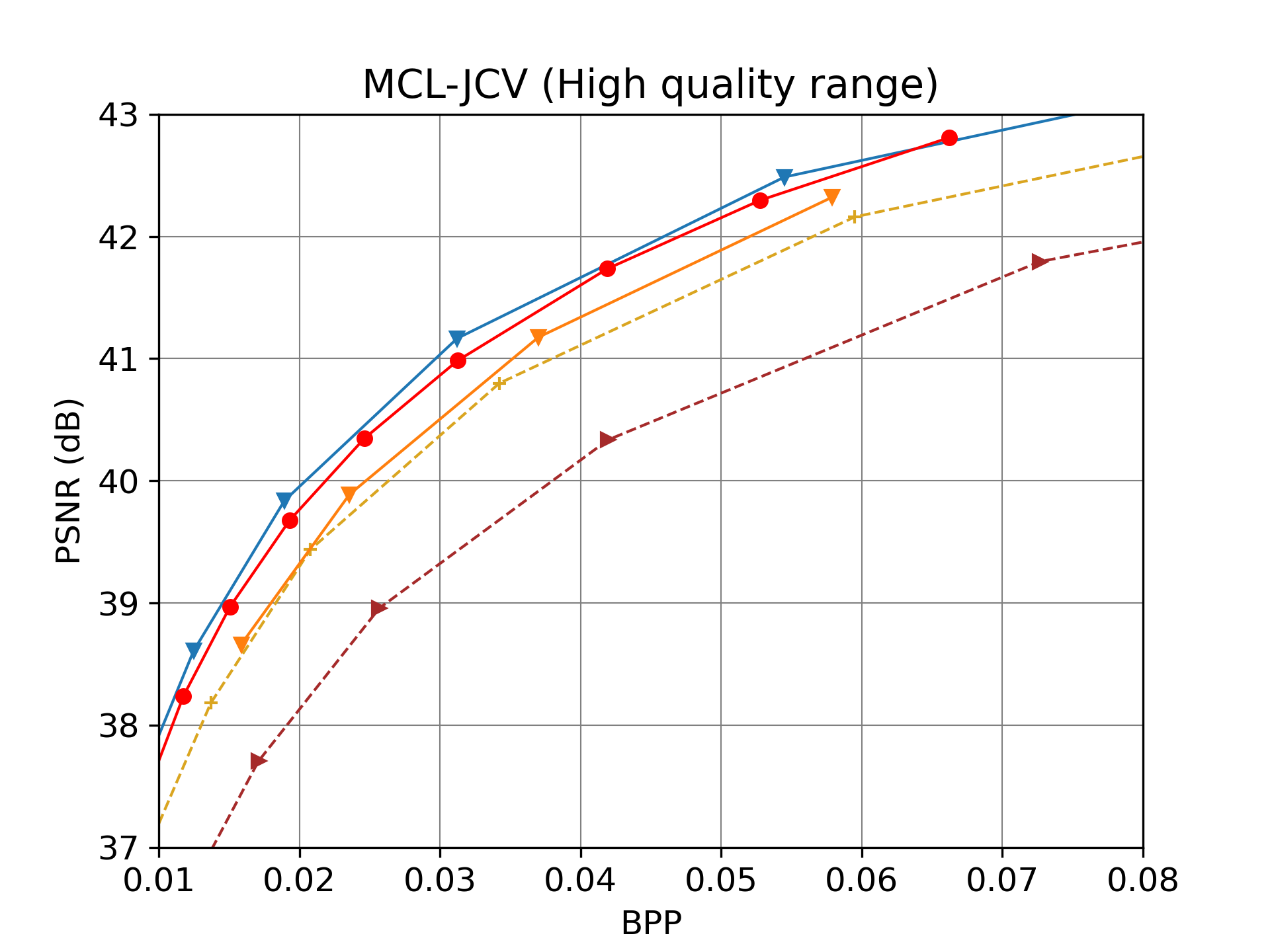}
	\endminipage
	
	\vspace{-4mm}
	\caption{Rate and distortion curves for HEVC E, UVG, and MCL-JCV datasets. Each row shows a dataset.
		From left to right the figures are overall quality range, relatively low quality range and relatively high quality range, respectively. The comparison is in YUV420 colorspace. All frames with intra-period = --1.  }
	\vspace{-4mm}
	\label{fig_yuv_psnr_allf_curve_part2}
\end{figure*}

In this document, we show the rate-distortion (RD) curves of all datasets with testing all frames and using intra period --1 setting. Fig. \ref{fig_yuv_psnr_allf_curve_part1} shows the results of HEVC B, C, and D datasets, and  Fig. \ref{fig_yuv_psnr_allf_curve_part2} shows those of HEVC E, UVG, and  MCL-JCV datasets. In these two figures, we also compare the relatively low and high quality regions, respectively. From these comparisons, we can see that previous SOTA NVC DCVC-DC \cite{li2023neural} has quite limited quality range. By contrast, our DCVC-FM has much wider range, and achieves the best RD performance over all previous codecs for many cases.

But as shown in Fig. \ref{fig_yuv_psnr_allf_curve_part2}, our DCVC-FM cannot surpass ECM on MCL-JCV dataset. MCL-JCV includes screen content videos. We find our codec is not good at screen content. %For example, our codec has 241.0\% bitrate increase over ECM-5.0 on \textit{videoSRC20} video. 
It is because our training dataset Vimeo is a natural content dataset. In addition, our codec also  performs worse on video with lots of noise. %Our codec has 110.7\%, 97.1\%, and 47.0\% bitrate increase over  ECM-5.0 on \textit{videoSRC06}, \textit{videoSRC15}, and \textit{videoSRC13}, respectively. 
Currently, it is quite hard for neural codec to code the random noise in source video as the probability of noise data is difficult to predict accurately. In the future, we will improve our codec on these videos.

\section{Smooth Quality Adjustment in Single Model}

\begin{figure*}[t]
	
	\minipage{0.33\textwidth}
	\includegraphics[width=1.07\linewidth,height=0.9\linewidth]{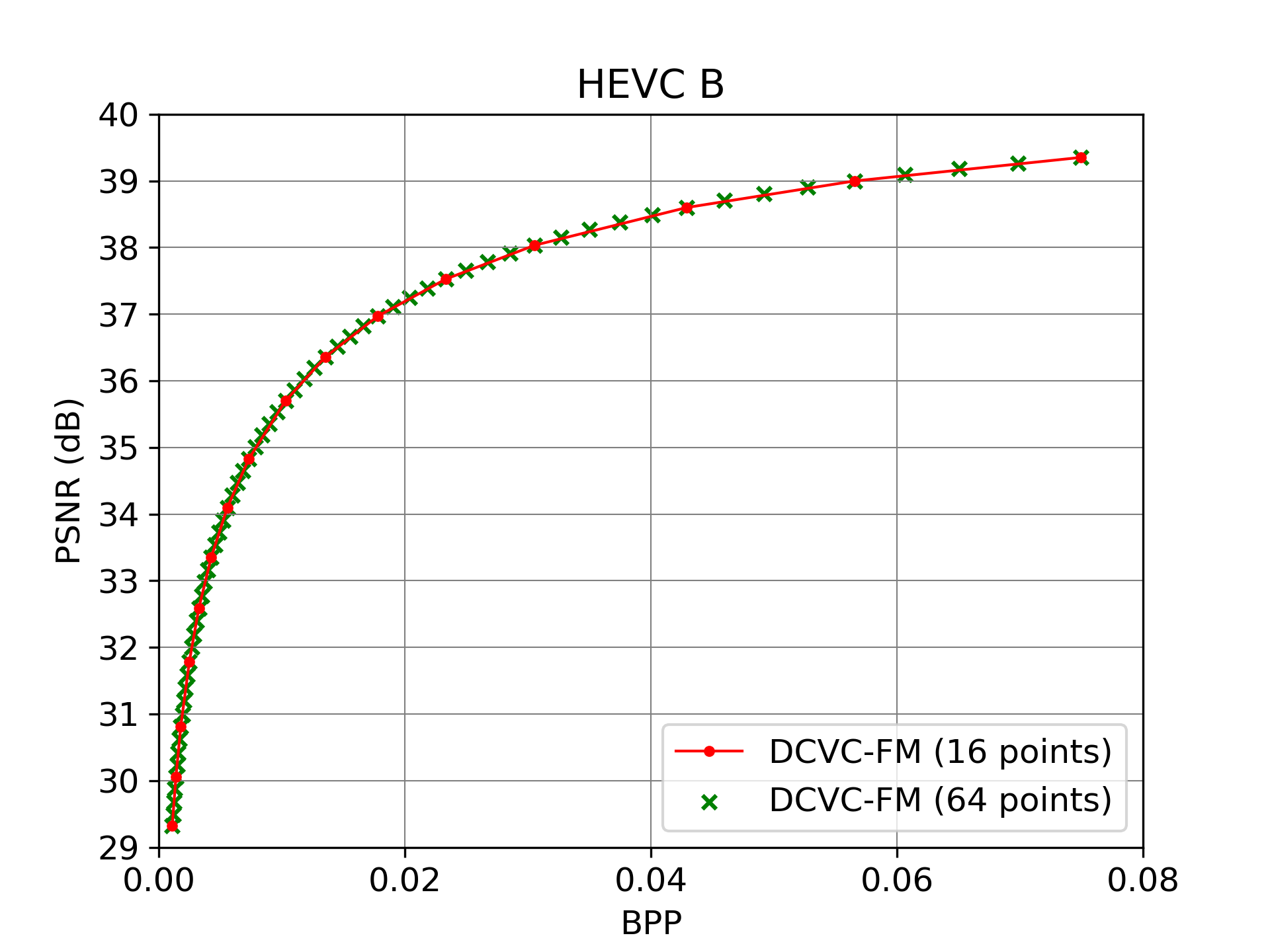}
	\endminipage
	\minipage{0.33\textwidth}
	\includegraphics[width=1.07\linewidth,height=0.9\linewidth]{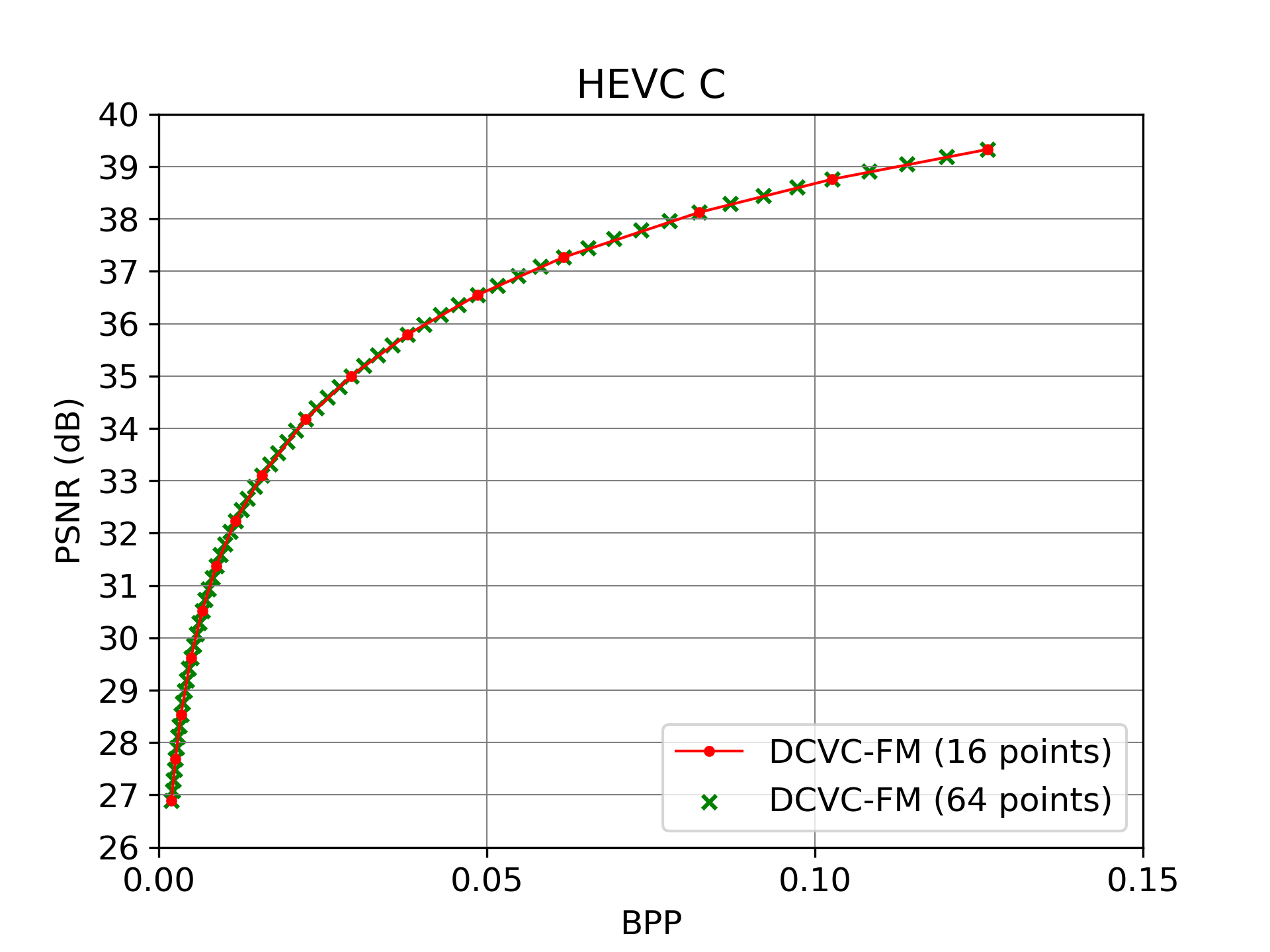}
	\endminipage
	\minipage{0.33\textwidth}%
	\includegraphics[width=1.07\linewidth,height=0.9\linewidth]{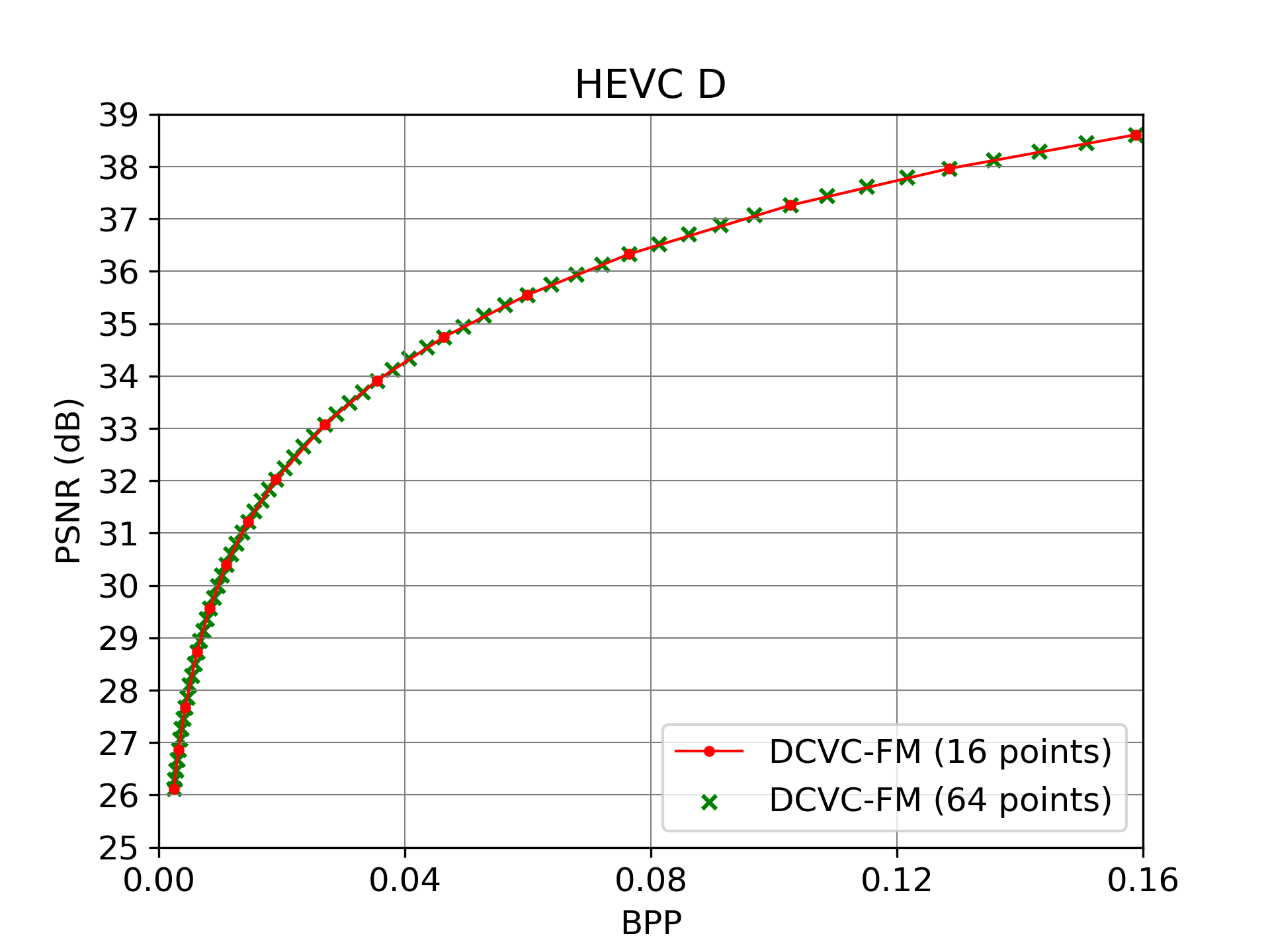}
	\endminipage
	
	\minipage{0.33\textwidth}
	\includegraphics[width=1.07\linewidth,height=0.9\linewidth]{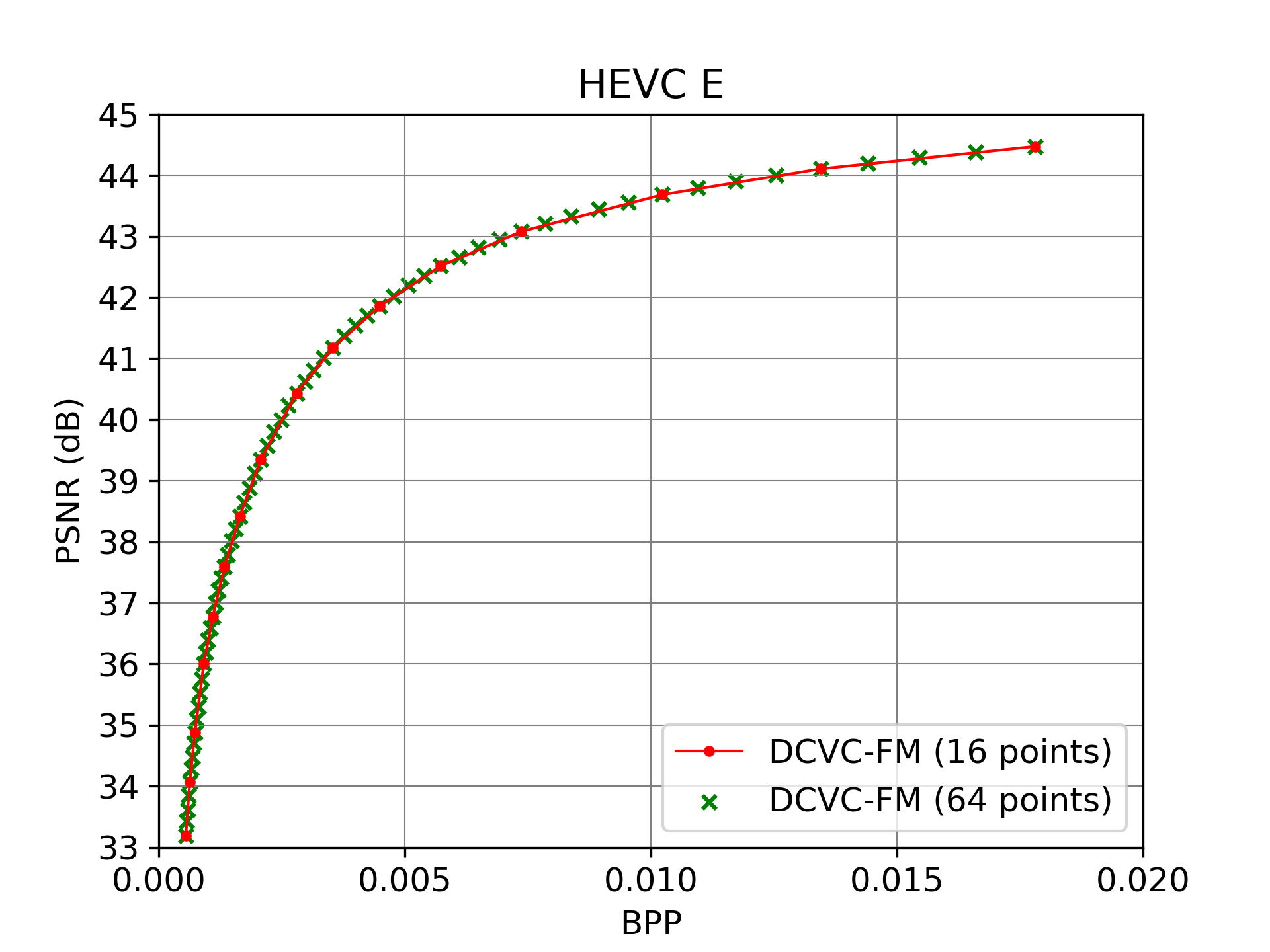}
	\endminipage
	\minipage{0.33\textwidth}
	\includegraphics[width=1.07\linewidth,height=0.9\linewidth]{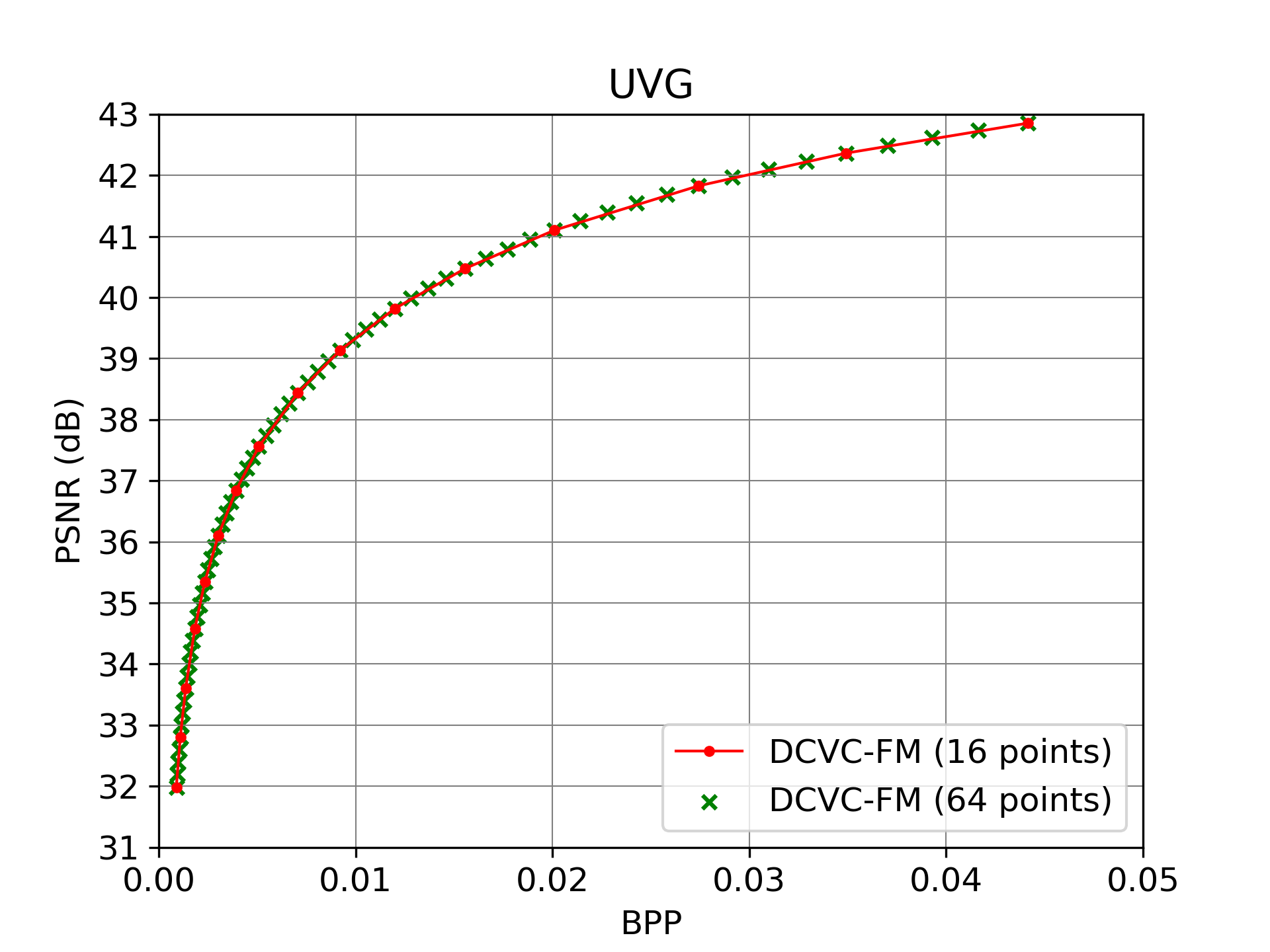}
	\endminipage
	\minipage{0.33\textwidth}%
	\includegraphics[width=1.07\linewidth,height=0.9\linewidth]{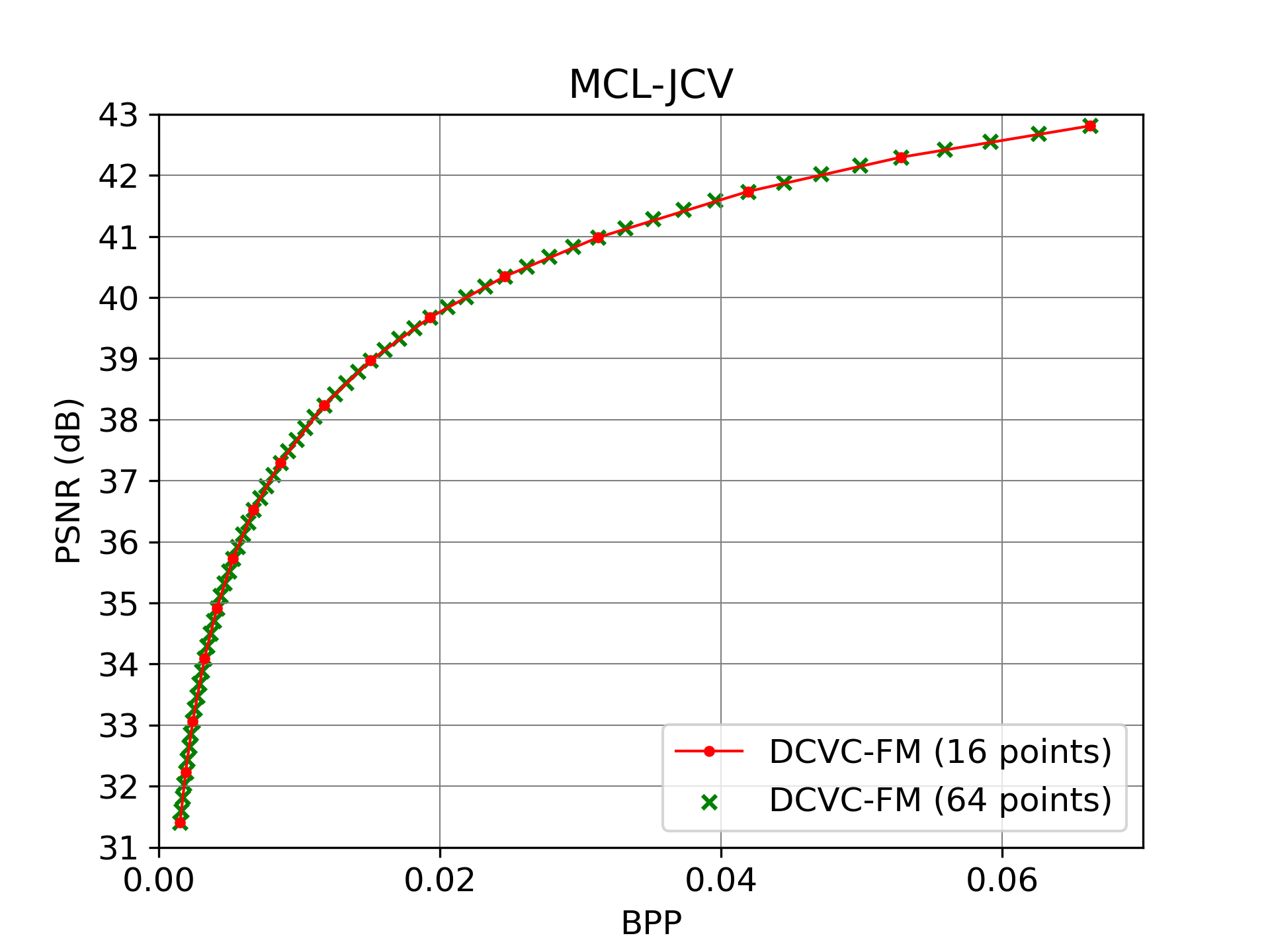}
	\endminipage

	\caption{Smooth quality adjustment in single model. Our DCVC-FM supports 64 different  quality levels. }
	\label{fig_fine_adjustment}
\end{figure*}

For convenience, we test 16 RD points when comparing our DCVC-FM with other codecs. Actually, our NVC can support 64 different quality levels in single model. We test all these RD points, as shown in Fig. \ref{fig_fine_adjustment}. From these figures, we can see that our DCVC-FM can achieve very smooth quality adjustment in single model, and there is no any outlier in the RD curves. This is also the prerequisite of achieving precise rate control.%, and we have demonstrated the capability of rate control in the main paper.

\section{Visual Comparison}

\begin{figure*}[t]
	\begin{center}
		\includegraphics[width=1\linewidth]{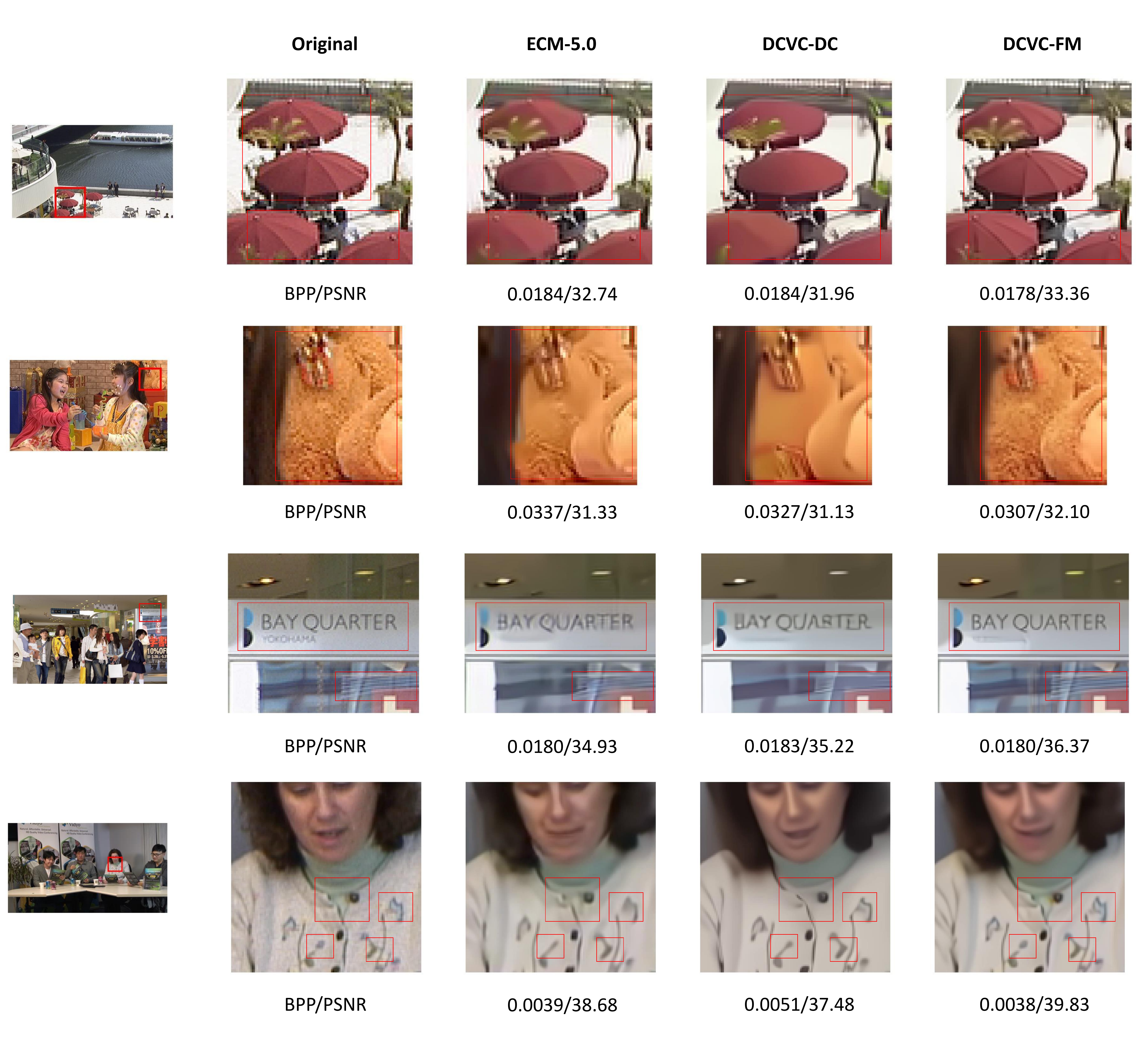}
	\end{center}
	\vspace{-0.5cm}
	\caption{Visual comparisons.	}
	\vspace{-3mm}
	\label{supp_visual_example}
\end{figure*}

In this section, we offer visual comparisons to illustrate the superior performance of our DCVC-FM. Four examples are presented in Fig. \ref{supp_visual_example}. These examples demonstrate that our DCVC-FM is capable of reconstructing textures with greater clarity, without incurring additional bitrate costs, when compared with   traditional codec ECM and previous SOTA NVC DCVC-DC.

\section{Traditional Codec Using B Frame Config}
Currently our neural codec focuses on low-delay scenario, so the traditional codec uses low-delay-B (LDB) setting for fair comparison  in the main paper. Actually, we also already tested the hierarchical-B (HieB) setting (the random-access config with intra-period = --1, where the low-delay requirement is broken) for ECM-5.0 under 96 frames, as shown in  Table \ref{tab_yuv_psnr_96f_hieb}.  The compression ratio gap between LDB and HieB is about 26\%, consistent with the number reported in JCTVC-K0279 (21\% on average for HEVC). Designing a neural codec which surpasses the best traditional codec in HieB setting will be future work.

\begin{table}[h]
	\centering
	\captionsetup{font=scriptsize}
	%\vspace{-0.3cm}
	\caption{ BD-Rate (\%) comparison (YUV420, 96 frames, intra-period = --1).
	}
	\vspace{-0.3cm}
	\scalebox{0.6}{
		\renewcommand{\arraystretch}{1.3}
		\small
		\begin{tabular}{ccccccccc}
			\toprule[1.0pt]
			& UVG    & MCL-JCV  & HEVC B & HEVC C & HEVC D   & HEVC E        & Average        \\ \hline
			
			VTM-17.0 (LDB)	    & 0.0      & 0.0    & 0.0       & 0.0    &  0.0     & 0.0       & 0.0            \\ \hline
			ECM-5.0	(LDB)       &--13.3	   &--16.4	&--14.6     &--15.6	 &--14.1  	&--12.6     &--14.4          \\ \hline
			ECM-5.0	(HieB)      &--40.3    &--40.5	&--42.3	    &--39.1	 &--39.3	  &--39.4     &--40.2          \\ \hline
			DCVC-FM (Low-delay)     &--25.4	   &--11.6	&--17.1	    &--24.4	 &--41.5	&--31.6		&--25.3          \\

			\bottomrule[1.0pt]
		\end{tabular}
	}
	%\vspace{-0.4cm}
	\label{tab_yuv_psnr_96f_hieb}
\end{table}

\end{appendices}

\end{document}

%% file: preamble.tex
\usepackage[dvipsnames]{xcolor}